\documentclass[10pt,twocolumn,letterpaper]{article}

\usepackage{iccv}
\usepackage{times}
\usepackage{epsfig}
\usepackage{graphicx}
\usepackage{amsmath}
\usepackage{amssymb}
\usepackage{bbm}
\usepackage[american]{babel}

\makeatletter
\newif\if@restonecol
\makeatother

\usepackage[linesnumbered,ruled,vlined]{algorithm2e}%[ruled,vlined]{
\usepackage{algpseudocode}
\usepackage{bm}% for bold 

\usepackage{multirow} 

\newcommand{\squishlist}{
	\begin{list}{$\bullet$}
		{ \setlength{\itemsep}{0pt}
			\setlength{\parsep}{2pt}
			\setlength{\topsep}{2pt}
			\setlength{\partopsep}{0pt}
			\setlength{\leftmargin}{1em}
			\setlength{\labelwidth}{1em}
			\setlength{\labelsep}{0.5em} } }
	\newcommand{\squishend}{
\end{list} }

\usepackage[pagebackref=true,breaklinks=true,colorlinks,bookmarks=false]{hyperref}

\iccvfinalcopy % *** Uncomment this line for the final submission

\def\iccvPaperID{3286} % *** Enter the ICCV Paper ID here

\ificcvfinal\fi
\begin{document}
	
	%%%%%%%%% TITLE
	\title{From Open Set to Closed Set: Counting Objects by Spatial Divide-and-Conquer\thanks{
			Haipeng Xiong and Hao Lu contributed equally. Zhiguo Cao is the corresponding author.
		}
	}
	
	\author{Haipeng Xiong$^\dagger$, Hao Lu$^\ddagger$, Chengxin Liu$^\dagger$, Liang Liu$^\dagger$, Zhiguo Cao$^\dagger$, Chunhua Shen$^\ddagger$\\
		$^\dagger$Huazhong University of Science and Technology, China\\
		$^\ddagger$The University of Adelaide, Australia\\
		{\tt\small \{hpxiong,zgcao\}@hust.edu.cn,  hao.lu@adelaide.edu.au }
		\and
		%Hao Lu\\
		%The University of Adelaide\\
		%{\tt\small hao.lu@adelaide.edu.au}
	}
	
	\maketitle
	\thispagestyle{empty}
	
	%%%%%%%%% ABSTRACT
	\begin{abstract}
		
		%Counting aims at predicting the number of individuals in the image, and its target, namely count value, varies in $[0,+\infty)$. It is naturally an open-set problem and commonly modelled in a regression manner. But in fact, the observed images and count values in datasets are limited, which means only a small closed set of counts can be learnt. However, counting is decomposable. When facing new image with more objects, we can continuously divide the image until the counts of sub-images are within the learnt closed set. Inspired from this idea, we propose a simple but effective Spatial Divide-and-Conquer Network (S-DCNet). The frontend of S-DCNet adopts a VGG16 encoder to extract feature maps and an Unet-like decoder to divide in the aspect of features. For the backend, a classifier for learning a closed set of counts and a division decider for merging division results are embedded. S-DCNet is trained end-to-end and presents the state-of-the-art performance on three crowd counting (ShanghaiTech, UCF\_CC\_50 and UCF-QNRF datasets) and an agricultural dataset (MTC dataset). Specifically, S-DCNet brings a $20.2\%$ boost in ShanghaiTech Part\_B, $20.9\%$ in UCF-QNRF  and $15.1\%$ in MTC datasets when compared to the previous best method.
		
		Visual counting, a task that predicts the number of objects from an image/video, is an open-set problem by nature, i.e., the number of population can vary in $[0,+\infty)$ in theory. However, the collected images and labeled count values are limited in reality, which means only a small closed set is observed. Existing methods typically model this task in a regression manner, while they are likely to suffer from an unseen scene with counts out of the scope of the closed set. In fact, counting is decomposable. A dense region can always be divided until sub-region counts are within the previously observed closed set. Inspired by this idea, we propose a simple but effective approach, Spatial Divide-and-Conquer Network (S-DCNet). 
		%%%%%%%%%%%%%%%
		% HL, introduce the high-level key idea of S-DCNet and its property, e.g., D&C on the feature map
		S-DCNet only learns from a closed set but can generalize well to open-set scenarios via S-DC. S-DCNet is also efficient. To avoid repeatedly computing sub-region convolutional features, S-DC is executed on the feature map instead of on the input image.
		%The frontend of S-DCNet adopts a VGG16 encoder to extract feature maps and an Unet-like decoder to divide in the aspect of features. For the backend, a classifier for learning a closed set of counts and a division decider for merging division results are embedded. 
		%%%%%%%%%%%%%%%
		S-DCNet achieves the state-of-the-art performance on three crowd counting datasets (ShanghaiTech, UCF\_CC\_50 and UCF-QNRF),  a vehicle counting dataset (TRANCOS) and a plant counting dataset (MTC). Compared to the previous best methods, S-DCNet brings a $20.2\%$ relative improvement on the ShanghaiTech Part\_B, $20.9\%$ on the UCF-QNRF, $22.5\%$ on the TRANCOS and $15.1\%$ on the MTC. Code has been made available at: \url{https://github.com/xhp-hust-2018-2011/S-DCNet}.
		
	\end{abstract}
	
	%%%%%%%%% BODY TEXT
	\section{Introduction}\label{sec:intro}
	
	The task of visual counting in Computer Vision is to infer the number of objects (people, cars, maize tassels, etc.) from an image/video. It has wide applications, such as automatic crowd management~\cite{UCFCC50_2013_CVPR,Compose_Loss_2018_ECCV,blobs_2018_ECCV,Zhang_2015_CVPR,MCNN_2016_CVPR}, traffic monitoring~\cite{TRANCOSdataset_IbPRIA2015,O2016Towards_CCNN}, and crop yield estimation~\cite{Fernandezgallego2018Wheat,Giuffrida2015Learning_global,Lu2017TasselNet}. Extensive attention has been received in recent years.
	
	\begin{figure}[t]
		\begin{center}
			\includegraphics[width=0.8\linewidth]{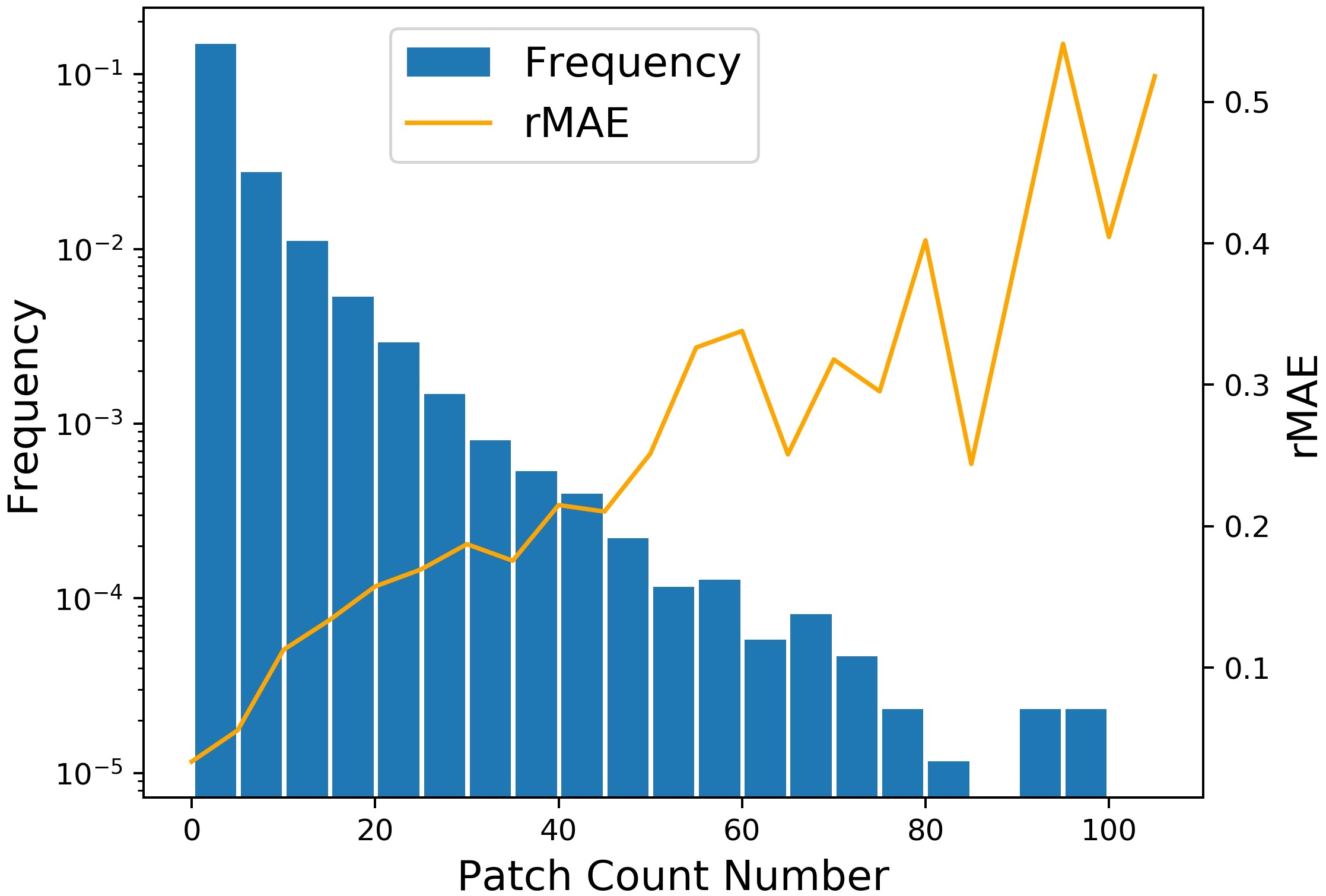}
		\end{center}
		\vspace{-10pt}
		\caption{ The histogram of count values of $64\times64$ local patches on the test set of ShanghaiTech Part\_A dataset~\cite{MCNN_2016_CVPR}. The orange curve denotes the relative mean absolute error (rMAE) of CSRNet~\cite{CSRNet_2018_CVPR} on local patches. }
		%It is worth mentioning that CSRNet is a typical density map regression method, which achieves the state-of-the-art counting performance in ShanghaiTech dataset.  }
		\label{fig:image_dis_rmae}
		\vspace{-10pt}
	\end{figure}
	
	Counting is an open-set problem by nature as a count value can range from $0$ to $+\infty$ in theory. It is thus typically modeled in a regression manner. Benefiting from the success of convolutional neural networks (CNNs), state-of-the-art deep counting networks
	%mainly have two paradigm: density map regression and local counts regression~\cite{Divide_grow_2018_CVPR,Count_ception_2017_ICCVW,DecideNet_2018_CVPR,Lu2017TasselNet,ICNN_2018_ECCV,DeepNegCor_2018_CVPR,CPCNN_2017_ICCV,tip2019divide,MCNN_2016_CVPR}. Most works~\cite{DecideNet_2018_CVPR,ICNN_2018_ECCV,CPCNN_2017_ICCV,MCNN_2016_CVPR} 
	often adopt a multi-branch architecture to enhance the feature robustness to dense regions~\cite{SwitchCNN_2017_CVPR,SANet_2018_ECCV,MCNN_2016_CVPR}. 
	%Some others~\cite{Divide_grow_2018_CVPR,DeepNegCor_2018_CVPR} draw inspirations from ensemble learning and integrate a set of networks to tackle different scenes. 
	However, the observed patterns in datasets are limited in practice, which means networks can only learn from a closed set. \emph{Are these counting networks still able to generate accurate predictions when the number of objects is out of the scope of the closed set?}
	Meanwhile, observed local counts exhibit a long-tailed distribution shown in Figure~\ref{fig:image_dis_rmae}. Extremely dense patches are rare while sparse patches take up the majority. As what can be observed, the relative mean absolute error (rMAE) increases significantly with increased local density. \emph{Is it necessary to set the working range of CNN-based regressors to the maximum count value observed, even with a majority of samples are sparse such that the regressor works poorly in this range?}
	
\begin{figure}[t]
		\begin{center}
			\includegraphics[width=0.8\linewidth]{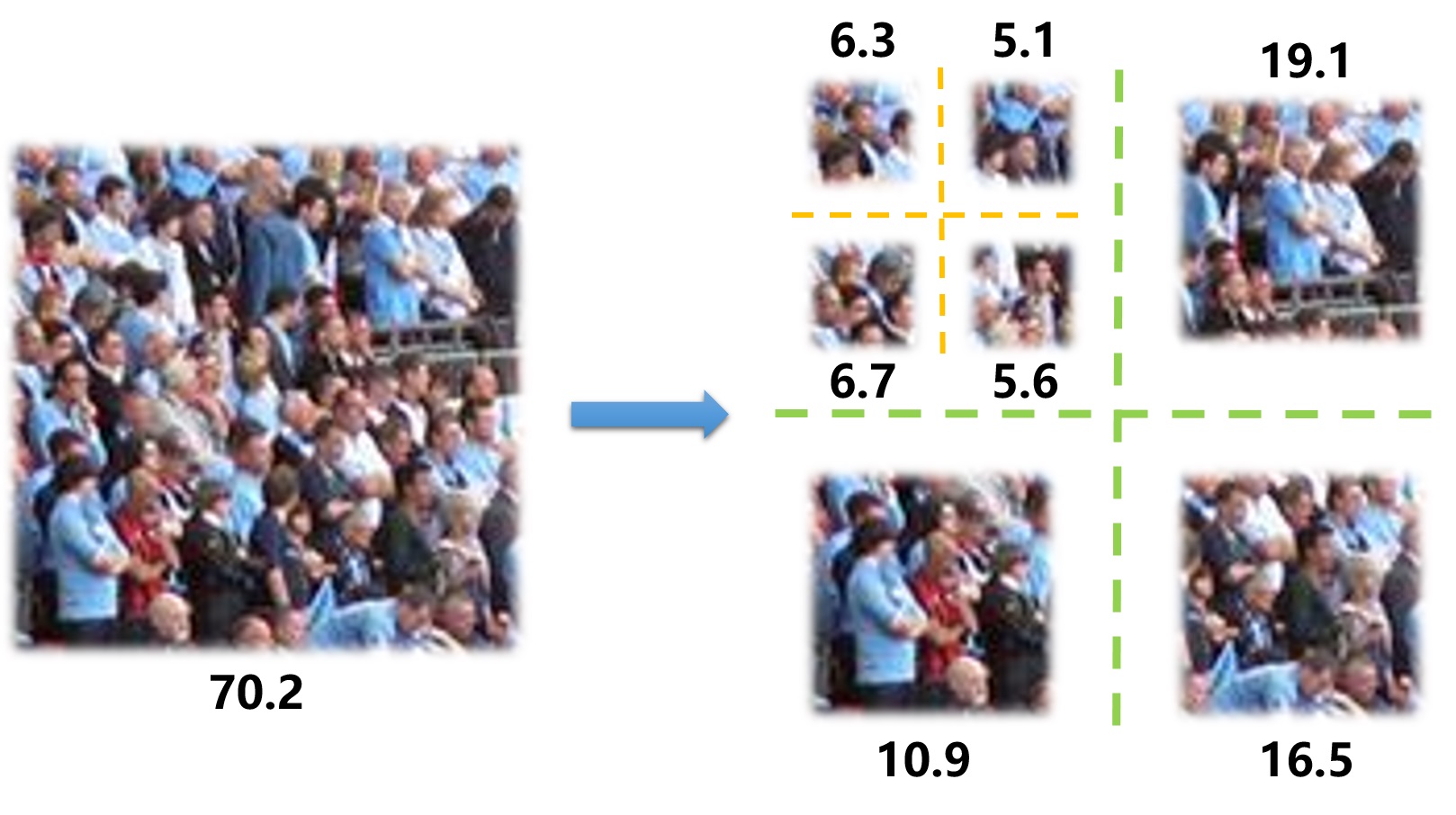}
		\end{center}
		\vspace{-10pt}
		\caption{An illustration of spatial divisions. Suppose that the closed set of counts is $[0, 20]$. In this example, dividing the image for one time is inadequate to ensure that all sub-region counts are within the closed set. For the top left sub-region, it needs a further division. }
		\label{fig:image_divide_example}
	\end{figure}

\begin{figure}[t]
		\begin{center}
			\includegraphics[width=0.9\linewidth]{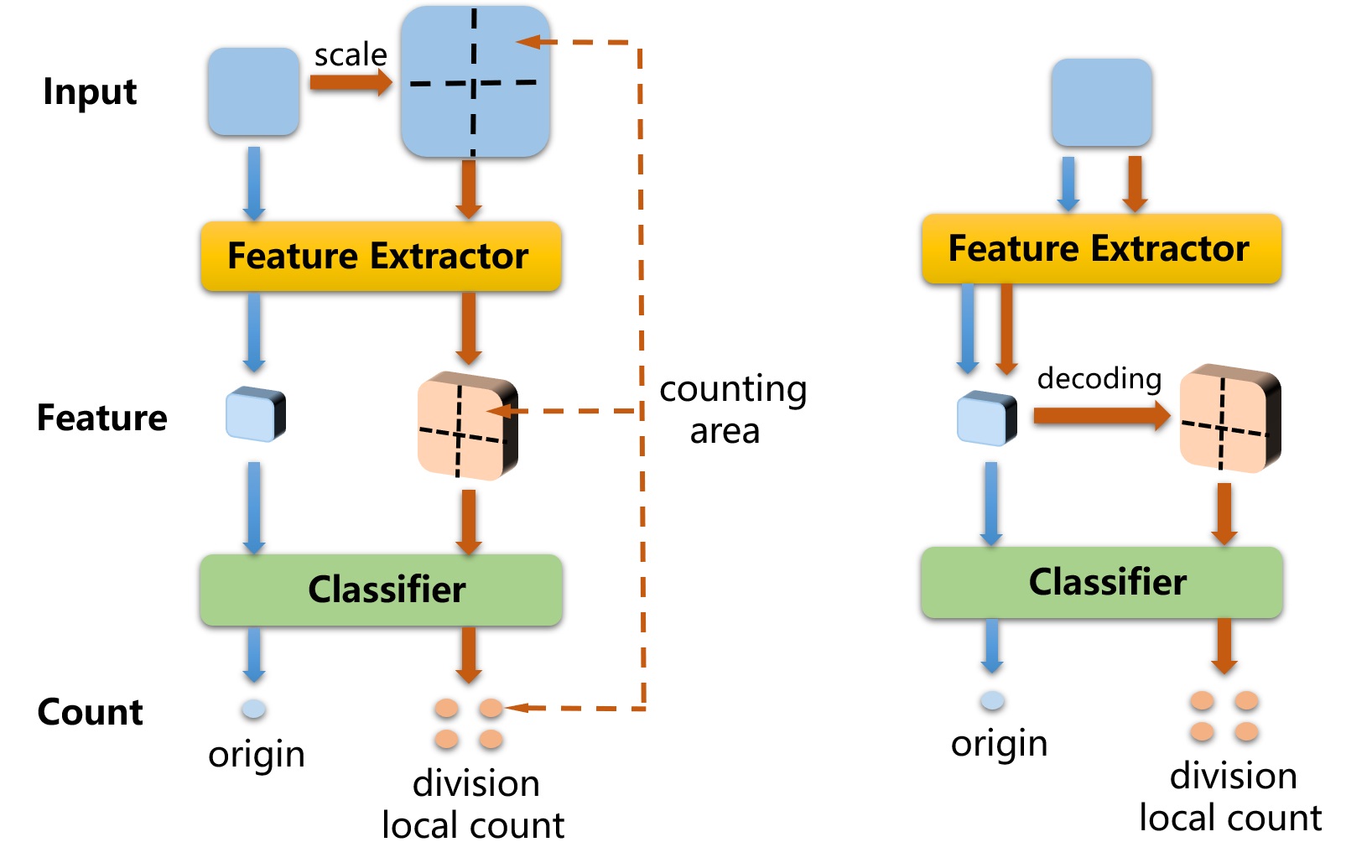}
		\end{center}
		\vspace{-10pt}
		\caption{ Spatial divisions on the input image (left) and the feature map (right). Spatially dividing the input image is straightforward. The image is upsampled and fed to the same network to infer counts of local areas. The orange dashed line is used to connect the local feature map, the local count and the sub-image. S-DC on the feature map avoids redundant computations and is achieved by upsampling, decoding and dividing the feature map of high resolution.}
		\label{fig:feature_divide}
		\vspace{-10pt}
	\end{figure}

	In fact, counting has an unique property---spatially decomposable. The above problem can be largely alleviated with the idea of spatial divide-and-conquer (S-DC). Suppose that a network has been trained to accurately predict a closed set of counts, say $0\sim20$. When facing an image with extremely dense objects, one can keep dividing the image into sub-images until all sub-region counts are less than $20$. Then the network can accurately count these sub-images and sum over all local counts to obtain the global image count. Figure~\ref{fig:image_divide_example} graphically depicts the idea of S-DC. %That is to say images with a large number of objects are composed of local sub-images with less objects.    
	A follow-up question is how to spatially divide the count. A naive way is to upsample the input image, divide it into sub-images and process sub-images with the same network. This way, however, is likely to blur the image and lead to exponentially-increased computation cost and memory consumption when repeatably extracting the feature map. Inspired by RoI pooling~\cite{girshick2015fast}, we show that it is feasible to achieve S-DC on the feature map, as conceptually illustrated in Figure~\ref{fig:feature_divide}. By decoding and upsampling the feature map, the later prediction layers can focus on the feature of local areas and predict sub-region counts accordingly.
	
	To realize the above idea, we propose a simple but effective Spatial Divide-and-Conquer Network (S-DCNet). S-DCNet learns from a closed set of count values but is able to generalize to open-set scenarios. Specifically, \mbox{S-DCNet} adopts a VGG16~\cite{Simonyan2014Very_VGG16}-based encoder and an UNet~\cite{Unet2015U}-like decoder to generate multi-resolution feature maps. All feature maps share the same counting predictor. Inspired by~\cite{li2018deep}, in contrast to the conventional density map regression, we discretize continuous count values into a set of intervals and design the counting predictor to be a classifier. Further, a division decider is designed to decide which sub-region should be divided and to merge different levels of sub-region counts into the global image count. We show through a controlled toy experiment that, even given a closed training set, S-DCNet effectively generalizes to the open test set. The effectiveness of \mbox{S-DCNet} is further demonstrated on three crowd counting datasets (ShanghaiTech~\cite{MCNN_2016_CVPR}, UCF\_CC\_50~\cite{UCFCC50_2013_CVPR} and UCF-QNRF~\cite{Compose_Loss_2018_ECCV}), a vehicle counting dataset (TRANCOS~\cite{TRANCOSdataset_IbPRIA2015}), and a plant counting dataset (MTC~\cite{Lu2017TasselNet}). Results show that S-DCNet indicates a clear advantage over other competitors and sets the new state-of-the-art across five datasets.
	
	The main contribution of this work is that we propose to transform open-set counting into a closed-set problem. We show through extensive experiments that a model learned in a closed set can effectively generalize to the open set with the idea of S-DC. %In particular, we present a novel deep counting network S-DCNet and demonstrate its effectiveness on a toy and five real-world datasets. %achieves state-of-the-art performance over five object counting benchmarks.
	
	%Initial works are formulated in the detection paradigm~\cite{} 
	% To summarize, the main contributions of this paper are:
	% \squishlist
	% \item A novel S-DCNet, which adopts spatial divide-and-conquer strategy to transfer counting  from open-set to close-set problem.
	
	% \item Our method reports the state-of-the-art results on  four challenging datasets (ShanghaiTech, UCF\_CC\_50, UCF-QNRF and MTC datasets) and brings huge improvements.
	% \squishend
	%!!!\emph{Say the motivation and the thoughts of Spatial DCNet here, not in the method. }

	%-------------------------------------------------------------------------
	\section{Related Work}
	Current CNN-based counting approaches are mainly built upon the framework of local regression. According to their regression targets, they can be categorized into two categories: density map regression and local count regression. We first review these two types of regression. Since S-DCNet learns to classify counts, some works that reformulate the regression problem are also discussed.
	
	\vspace{-10pt}
	\paragraph{Density Map Regression}
	The concept of density map was introduced in~\cite{vlaz2010denlearn}. The density map contains the spatial distribution of objects, thus can be smoothly regressed. Zhang \etal~\cite{Zhang_2015_CVPR} first adopted a CNN to regress local density maps. Then almost all subsequent counting networks followed this idea. Among them, a typical network architecture is multi-branch. MCNN~\cite{MCNN_2016_CVPR} and Switching-CNN~\cite{SwitchCNN_2017_CVPR} used three columns of CNNs with varying receptive fields to depict objects of different scales. SANet~\cite{SANet_2018_ECCV} adopted Inception~\cite{GoogleNet_2015_CVPR}-liked modules to integrate extra branches. 
	CP-CNN~\cite{CPCNN_2017_ICCV} added two extra density-level prediction branches to combine global and local contextual information. ACSCP~\cite{ACSCP_2018_CVPR} inserted a child branch to match cross-scale consistency and an adversarial branch to attenuate the blurring effect of the density map. ic-CNN~\cite{ICNN_2018_ECCV} incorporated two branches to generate high-quality density maps in a coarse-to-fine manner. IG-CNN~\cite{Divide_grow_2018_CVPR} and D-ConvNet~\cite{DeepNegCor_2018_CVPR} drew inspirations from ensemble learning and trained a series of networks or regressors to tackle different scenes. DecideNet~\cite{DecideNet_2018_CVPR} attempted to selectively fuse the results of density map estimation and object detection for different scenes. Unlike multi-branch approaches, Idrees \etal~\cite{Compose_Loss_2018_ECCV} employed a composition loss and simultaneously solved several counting-related tasks to assist counting. CSRNet~\cite{CSRNet_2018_CVPR} benefited from dilated convolution which effectively expanded the receptive field to capture contextual information.
	
	Existing deep counting networks aim to generate high-quality density maps. However, density maps are actually in the open set as well. Detailed discussion of the open set problem in density maps is provided in the Supplement.%For a single point, different kernel sizes lead to different density values. When multiple objects exist and are close, density patterns are even much diverse. Since observed samples are limited, density maps are certainly in an open set.
	% Note that the density map is a form of dense regression target. Downsampled feature maps usually need to be upsampled, which introduces extra model parameters and computation cost in the decoding process. Local count regression, however, avoids the decoding process.
	
	%while the image samples used for training CNNs are from a finite subset of real world, and highly congested scenes are rare. These methods typically lead to under-estimates in high-density areas, as suggested by~\cite{CPCNN_2017_ICCV} and~\cite{liu2018}.
	
	\vspace{-10pt}
	\paragraph{Local Count Regression}
	Local count regression directly predicts count values of local image patches. This idea first appeared in~\cite{chen2012feature} where a multi-output regression model was used to regress region-wise local counts simultaneously.~\cite{Count_ception_2017_ICCVW} and~\cite{Lu2017TasselNet} introduced such an idea into deep counting. Local patches were first densely sampled in a sliding-window manner with overlaps, and a local count was then assigned to each patch by the network. Inferred redundant local counts were finally normalized and fused to the global count. Stahl \etal~\cite{tip2019divide} regressed the counts for object proposals generated by Selective Search~\cite{Uijlings2013Selective} and combined local counts using an inclusion-exclusion principle. Inspired by subitizing, the ability for a human to quickly counting a few objects at a glance, Chattopadhyay \etal~\cite{Count_everyday_2017_CVPR} transferred their focus to the problem of counting objects in everyday scenes. The main challenge thus became large intra-class variances rather than the occlusions and perspective distortions in crowded scenes.
	
	%They get inspiration from subitizing, which is the ability of human to quickly count little number at a glance, and sequentially subitize local patches in a image with CNN to obtain image count.
	
	While some above methods~\cite{Count_everyday_2017_CVPR,tip2019divide} also leverage the idea of spatial divisions, they still regress the open-set counts. Although local-region patterns are easier to be modelled than the whole image, the observed local patches are still limited. Since only finite local patterns (a closed set) can be observed, new scenes in reality have a high probability including objects out of the range (an open set). Moreover, dense regions with large count values are rare (Figure~\ref{fig:image_dis_rmae}) and the networks may suffer from sample imbalance. In this paper, we show that a counting network is able to learn from a closed set with a certain range of counts, say $0\sim20$, and then generalizes to an open set (including counts $>20$) via S-DC. 
	
	%We argue that it is effective to transfer the problem of counting dense patches to count sparser sub-patches via spatial divide-and-conquer. Based on this idea, we propose S-DCNet, which learns to count a closed set of local count values and generalizes to count the open set of counts in real life through spatial divide-and-conquer in the aspect of feature maps. 
	
	\vspace{-10pt}
	\paragraph{Beyond Naive Regression}
	
	Regression is a natural way to estimating continuous variables, such as age and depth. However, some literatures suggest that regression is encouraged to be reformulated as an ordinal regression problem or a classification problem, which enhances performance and benefits optimization~\cite{Cumulative_2013_CVPR,Ordinal_depth_2018_CVPR,li2018deep,Ordinal_age_2016_CVPR}. Ordinal regression is usually implemented by modifying well-studied classification algorithms and has been applied to the problem of age estimation~\cite{Ordinal_age_2016_CVPR} and monocular depth prediction~\cite{Ordinal_depth_2018_CVPR}. Li~\etal~\cite{li2018deep} further showed that directly reformulating regression to classification was also a good choice. Since count values share a similar property like age and depth, it motivates us to follow such a reformulation. In this work, S-DCNet follows~\cite{li2018deep} to discretize local counts and classify count intervals. Indeed, we observe in experiments that classification with S-DC works better than direct regression.
	
	\section{Spatial Divide-and-Conquer Network}
	
	In this section, we describe the transformation from quantity to interval which transfers count values into a closed set. We also explain in detail our proposed S-DCNet.
	%(talk about the motivation and principle in detail in introduction! Not here! )
	% As mentioned above, we can just learn to count up to a limited number of objects (a closed set). When meeting much more dense images, we can divide the counting area to sparse local areas that we can count. Rather than directly cropping the input, we try to divide feature map to meet this need. 
	
	\subsection{From Quantity to Interval }
	%here should justify the output classmap as O_i, and transform it to count map C_i !! 
	Instead of regressing an open set of count values, we follow~\cite{li2018deep} to discretize local counts and classify count intervals. %It is necessary to describe the transformation between quantity and category at first. 
	Specifically, we define an interval partition of $[0,+\infty)$ as $\{0\}$, $(0,C_1]$, $(C_2,C_{3}]$, ... , $(C_{M-1},C_{M}]$ and $(C_{M},+\infty)$. These $M+1$ sub-intervals are labeled to the $0$-th to the $M$-th classes, respectively. For example, if a count value is within $(C_2,C_{3}]$, it is labeled as the $2$-th class. In practice, $C_{M}$ should be not greater than the max local count observed in the training set.
	%When a local count is classified into the interval of $(C_{M},+\infty)$, S-DC will be applied to the corresponding region.
	
	The median of each sub-interval is adopted when recovering the count from the interval. Notice that, for the last sub-interval $(C_{M},+\infty]$, $C_{M}$ will be used as the count value if a region is classified into this interval. It is clear that \emph{adopting $C_{M}$ for the last class will cause a systematic error}, but the error can be mitigated via S-DC as what we will show in experiments.

	\begin{figure*}[t]
		\begin{center}
			\includegraphics[width=0.8\linewidth]{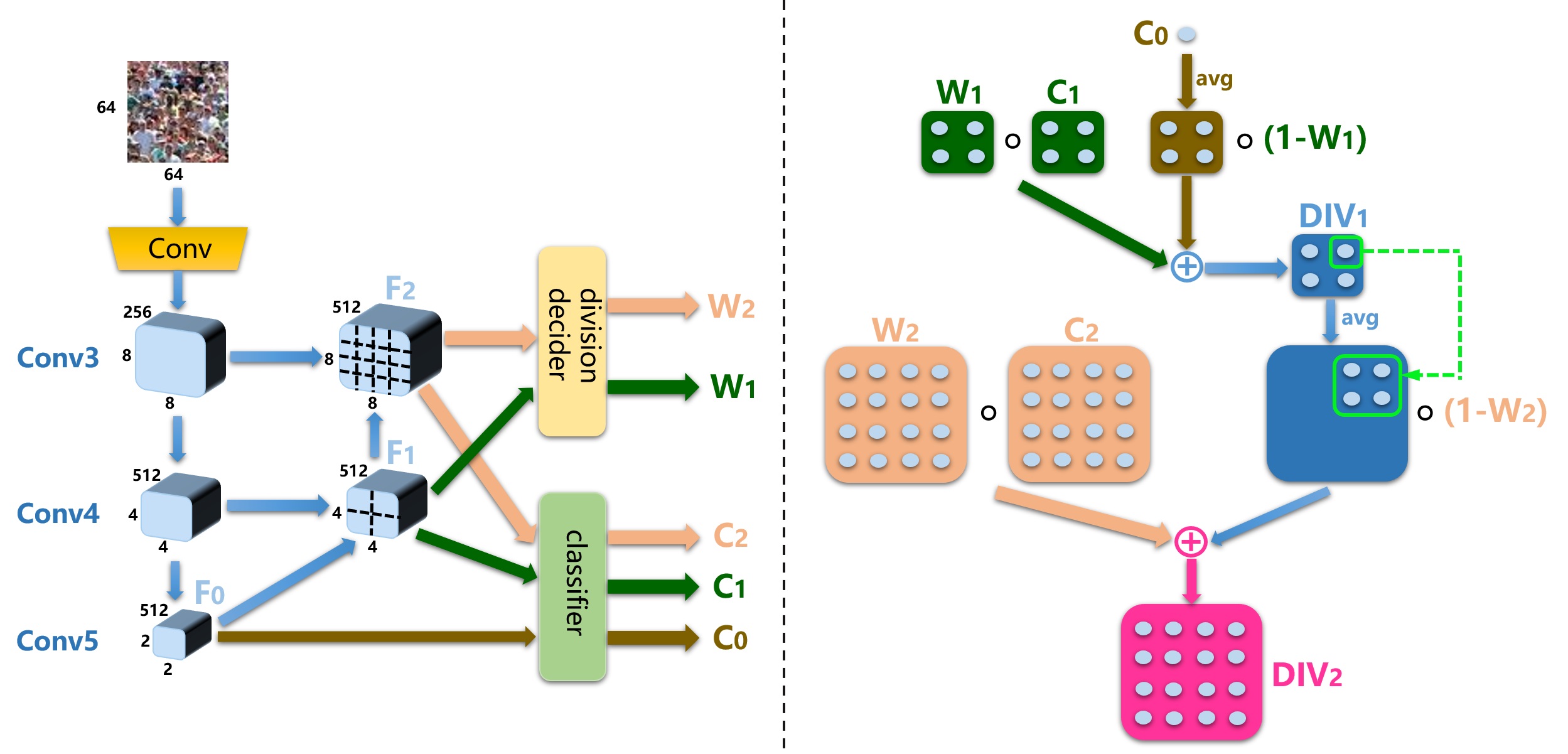}
		\end{center}
		\vspace{-10pt}
		\caption{ The architecture of S-DCNet (left) and a two-stage S-DC process (right). S-DCNet adopts all convolutional layers in VGG16~\cite{Simonyan2014Very_VGG16} while the first two convolutional blocks are simplified as $Conv$ in the figure. An UNet~\cite{Unet2015U}-like decoder is employed to upsample and divide the feature map as per Figure~\ref{fig:feature_divide}. A shared classifier and a division decider receive divided feature maps, and respectively, generate division counts $C_i$s and division masks $W_i$s, for $i=1,2,...$ After obtaining these results, $C_i$ and $W_i$ are merged to the $i$-th division count $DIV_i$ shown in the right sub-figure. Specially, we average each count of low resolution into the corresponding $2\times2$ area of high resolution before merging ($avg$ shown in the figure). ``$\circ$" denotes the Hadamard product. Note that, the $64\times64$ local patch is only used as an example for readers to understand the pipeline of S-DCNet. Since S-DCNet is a fully convolutional network, it can process images of arbitrary sizes $M\times N$ and return $DIV_2$s of size $\frac{M}{64}\times \frac{N}{64}$. The structures for the classifier and the division decider are presented in Table~\ref{tab:classifer_div_decider}.}
		\label{fig:SDCNet_architecture}
		\vspace{-10pt}
	\end{figure*}
	
	\begin{table}\footnotesize
		\begin{center}
			\begin{tabular}{|c|c|}
				\hline
				classifier & division decider\\
				\hline\hline
				
				$2\times2$ AvgPool, s $2$&  $2\times2$ AvgPool, s $2$ \\
				$1\times1$ Conv, $512$, s $1$ & $1\times1$ Conv, $512$, s $1$ \\
				$1\times1$ Conv, $class~num$, s $1$ & $1\times1$ Conv, $1$, s $1$\\
				$-$&Sigmoid\\
				
				\hline
			\end{tabular}
		\end{center}
		\caption{The architecture of \textit{classifier} and \textit{division decider}. $AvgPool$ denotes average pooling. Convolutional layers are defined in the format:  $Conv~size \times size$,  $output~channel$, $s~stride$. Each convolutional layer is followed by a ReLU function except the last layer. In particular, a \textit{sigmoid} function is employed at the end of \textit{division decider} to generate soft division masks.}
		\label{tab:classifer_div_decider}
	\end{table}

	\subsection{Single-Stage Spatial Divide-and-Conquer}  \label{sec:S-DCNet_Arc}   
	As shown in Figure~\ref{fig:SDCNet_architecture}, S-DCNet includes a VGG16~\cite{Simonyan2014Very_VGG16} feature encoder, an UNet~\cite{Unet2015U}-like decoder, a count-interval classifier and a division decider. The structure of the classifier and the division decider are shown in Table~\ref{tab:classifer_div_decider}. Notice that, the first average pooling layer in the classifier has a stride of $2$, so the final prediction has an output stride of $64$.
	
	The feature encoder removes fully-connected layers from the pre-trained VGG16. Suppose that the input patch is of size $64\times64$. Given the feature map $F_0$ (extracted from the Conv5 layer) with $\frac{1}{32}$ resolution of the input image, the classifier predicts the class label of the count interval $CLS_0$ conditioned on $F_0$. The local count $C_0$, which denotes the count value of the $64\times64$ input patch, can be recovered from $CLS_0$. Note that $C_0$ is the local count without S-DC, which is also the final output of previous approaches~\cite{Count_everyday_2017_CVPR,Count_ception_2017_ICCVW,Lu2017TasselNet}.  
	
	We execute the first-stage S-DC on the fused feature map $F_1$. $F_1$ is divided and sent to the shared classifier to produce the division count $C_1\in\mathbb{R}^{2\times2}$. Concretely, $F_0$ is upsampled by $\times2$ in an UNet-like manner to $F_1$. Given $F_1$, the classifier fetches the local features that correspond to spatially divided sub-regions, and predicts the first-level division counts $C_1$. Each of the $2\times2$ elements in $C_1$ denotes a sub-count of the corresponding $32\times32$ sub-region.
	
	With local counts $C_0$ and $C_1$, the next question is to decide where to divide. We learn such decisions with another network module, division decider, as depicted in the right part of Figure~\ref{fig:SDCNet_architecture}. At the first stage of S-DC, the division decider generates a soft division mask $W_1$ of the same size as $C_1$ conditioned on $F_1$ such that for any $w\in W_1, w\in[0,1]$. $w=0$ means no division is required at this position, and the value in $C_0$ is used. $w=1$ implies that here the initial prediction should be replaced with the division count in $C_1$. Since $W_1$ and $C_1$ are both $2$ times larger than $C_0$, $C_0$ is upsampled by $\times2$ to $\hat{C_0}$, and the count is averaged into the $2\times2$ local area in $\hat{C_0}$. The first-stage division result $DIV_1$ can thus be computed as 
	\begin{equation}\label{merge_45}
	DIV_1 = (\mathbbm{1}-W_1)\circ avg(C_0) + W_1 \circ C_1\,,
	\end{equation}
	where $\mathbbm{1}$ denotes a matrix filled with $1$ and is with the same size of $W_1$. ``$\circ$" denotes the Hadamard product. $avg$ is an averaging re-distribution operator (equally dividing a count value into a $2\times2$ region).
	
	\iffalse
	They are defined as eq.~\ref{Cross_Entropy_Loss} and ~\ref{L1Loss} separately.
	
	\begin{equation}\label{Cross_Entropy_Loss}
	DIV = (1-w_0)\times avg(C_0) + w_0 \times C_1 
	\end{equation}
	
	\begin{equation}\label{L1Loss}
	l1loss = |pre-gt| 
	\end{equation}
	\fi
	
	\begin{algorithm}[!t] \small
		\caption{ Multi-Stage S-DC}
		\label{alg:Overall algorith}
		\LinesNumbered
		\KwIn{Image $I$ and division time $N$}%, ground truth label $GT$ }
		\KwOut{Image count $C$}
		
		Extract $F_0$ from $I$\;
		
		Generate $CLS_0$ given $F_0$ with the classifier, and recover $C_0$ from $CLS_0$\;%as:
		%$CLS_0 = classifier(F_0)$, 
		%$C_0 = cls2num(CLS_0)$\;
		
		% Compute $L_c$ loss for $CLS_0$\;
		
		Initialize $DIV_0=C_0$\;
		
		% \For{$i=1;i \le N;i = i+1$}
		\For{$i\leftarrow 1$ \textbf{to} $N$}
		{
			Decode $F_{i-1}$ to $F_i$\;
			
			Process $F_i$ with the classifier and the division decider to obtain $CLS_i$ and the division mask $W_i$\;
			
			% 	Compute $L_c$ loss for $CLS_i$ and transforming $CLS_i$ to $C_i$\;
			Recover $C_i$ from $CLS_i$\;
			
			Update $DIV_i$ as per Eq.~\ref{merge_12345} \;  	
		}
		
		% 	Compute $L_1$ loss for $DIV_N$ and 
		Integrate over $DIV_N$ to obtain the image count $C$\;

		% 	\textbf{final} \;
		\textbf{return} $C$
	\end{algorithm}
	
	\subsection{Multi-Stage Spatial Divide-and-Conquer}
	
	S-DCNet can execute multi-stage S-DC by further decoding, dividing the feature map until reaching the output of the first convolutional block. In this sense, the maximum division time is $4$ in VGG16 (actually we show later in experiments that a two-stage division is adequate to guarantee satisfactory performance). In multi-stage S-DC, $DIV_i$ ($i\ge2$) is merged in a recursive manner as 
	\begin{equation}\label{merge_12345}
	DIV_i = (\mathbbm{1}-W_i)\circ avg(DIV_{i-1}) + W_i \circ C_i\,.
	\end{equation}
	
	% HL, place this part to the implementation details 
	%\subsection{Loss Function}
	We employ two types of standard loss functions to train S-DCNet: several cross-entropy losses $L_C^i$s that correspond to different classification outputs $CLS_i$s, and a $\ell_1$ loss $L_R^N$ for the final division output $DIV_N$ ($N$ denotes the division time). S-DCNet is learned in a multi-task manner where the overall loss $L$ is a summation of all losses, i.e., \mbox{$L=\sum_{i=0}^NL_C^i+L_R^N$}. Note that, $L_R^N$ is essential to provide an implicit supervision signal for learning $W_i$s. Multi-stage S-DCNet is summarized in Algorithm~\ref{alg:Overall algorith}.
	%for readers to better grasp the idea of our method.
	
	\section{Open Set or Closed Set? A Toy-Level Justification}
	%\paragraph{$(1)$ Open-set \& Closed-set}
	As aforementioned, counting is an open-set problem while the model is learned in a closed set. \emph{Can a closed-set counting model really generalize to open-set scenarios?} Here we show through a controlled toy experiment that, the answer is \textit{no}. %Actually, even in the most simple univariate regression, a well-trained regressor may fail when predicting outside the scope of the training set. We wonder whether such a phenomenon also appears in object counting. %In other words,~\emph{whether CNN-based regression approaches can generalize from closed-set learning to open-set prediction?} We notice that Existing datasets are always limited (closed sets), while the dense scenarios are such rare for adequate evaluation. 
	Inspired by~\cite{vlaz2010denlearn}, we synthesize a cell counting dataset to explore the counting performance outside a closed training set.
	
	\iffalse
	\begin{figure}[t]
		\begin{center}
			\includegraphics[width=0.8\linewidth]{toy_regression.jpg}
		\end{center}
		\vspace{-10pt}
		\caption{An example of univariate regression with $y=x^{3}$. We adopt a three-layer multi-layer perception (MLP) with $10$ hidden units as the regressor. Training samples are within the range of $[0,10]$. It can be observed the regressor performs precisely in the training range while fails outside such a range. }
		\label{fig:toy_regression}
		\vspace{-10pt}
	\end{figure}
	\fi
	
	\vspace{-10pt}
	\paragraph{Synthesized Cell Counting Dataset}
	We first generate $500$ $256\times256$ images with $64\times64$ sub-regions containing only $0\sim10$ cells to construct the training set (a closed set). To generate an open testing set, we further synthesize $500$ images with sub-region counts evenly distributed in the range of $[0,20]$. 
	%We do not consider local counts larger than $20$, because severe overlaps between cells will make cells impossible to be identified.%(when local counts greater than $20$, one can not count the cells with such severe overlap). 
	
	\vspace{-10pt}
	\paragraph{Baselines and Protocols}
	We adopt three approaches for comparisons, they are: 
	\textbf{i}) a regression baseline with pretrained VGG16 as the backbone and the \textit{classifier} module used in S-DCNet as the backend except that the output channel is modified to $1$. $\ell_1$ loss is used. This approach directly regresses the open-set counts;
	\textbf{ii}) a classification baseline with the same VGG16 and the \textit{classifier} settings as S-DCNet, without S-DC;
	\textbf{iii}) our proposed S-DCNet, which learns from a closed set but adapts to the open set via S-DC.  
	
	Regarding the discretization of count intervals, we choose $0.5$ as the step because cells can be partially presented in local patches. As a consequence, we have a partition of $\{0\}$, $(0.0.5]$,$(0.5,1]$, ... ,$(9.5,10]$ and $(10,+\infty)$. All approaches are trained with standard stochastic gradient descent (SGD). The learning rate is initially set to 0.001 and is decreased by $\times 10$ when the training error stagnates.
	
\begin{figure}[t]
		\begin{center}
			\includegraphics[width=\linewidth]{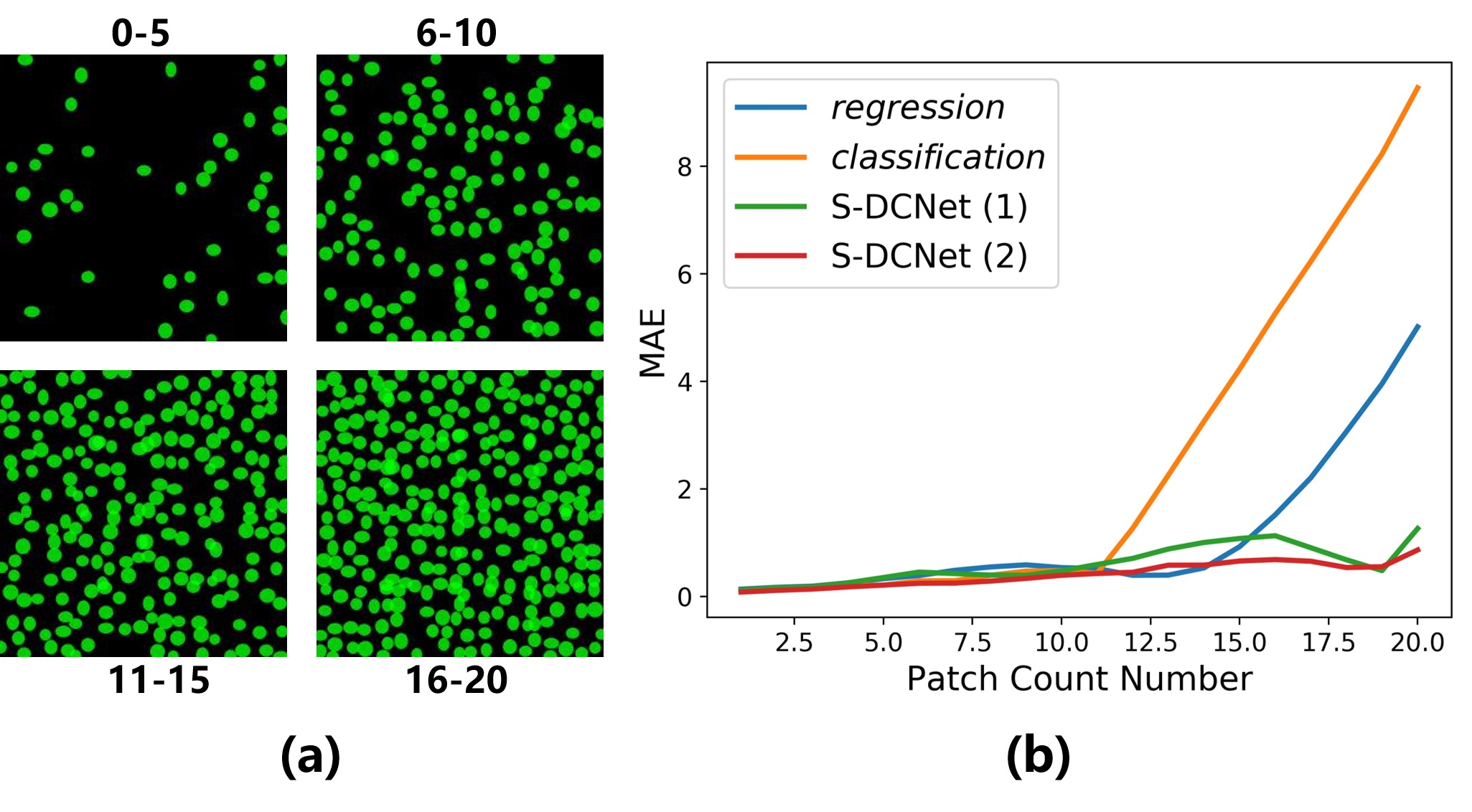}
		\end{center}
		\vspace{-10pt}
		\caption{A toy-level justification. \textbf{(a)} Some $256\times256$ images in the simulated cell counting dataset. The numbers denote the range of local counts of $64\times64$ sub-regions. \textbf{(b)} The mean absolute error (MAE) of different methods versus $64\times64$ sub-region counts. 
			%$regression$ and $classification$ denote just adopt the encoder part of S-DCNet for  regression or classification. 
			S-DCNet(N) means $N$-stage S-DCNet.}
		\label{fig:simulated_d_t}
		\vspace{-10pt}
	\end{figure}

	\vspace{-10pt}
	\paragraph{Observations}
	According to Figure~\ref{fig:simulated_d_t}, it can be observed that both regression and classification baselines work well in the range of the closed set ($0\sim10$), but the counting error increases rapidly when counts are larger than $10$. This suggests a conventional counting model learned in a closed set cannot generalize to the open set. However, S-DCNet can achieve accurate predictions even on the open set, which confirms the advantage of S-DC.
	% xhp: here may be not seriouly to say "linearly increases", can not prove it is a line

	\section{Experiments on Real-World Datasets}
	
	Extensive experiments are further conducted to demonstrate the effectiveness of S-DCNet on real-world datasets. We first describe some essential implementation details. After that, an ablation study is conducted on the ShanghaiTech Part\_A~\cite{MCNN_2016_CVPR} dataset to highlight the benefit of S-DC. Finally, we compare S-DCNet against current state-of-the-art methods on five public datasets. Mean Absolute Error (MAE) and Root Mean Squared Error (MSE) are used as the evaluation metrics following~\cite{MCNN_2016_CVPR}.
	
	% Mean Absolute Error ($MAE$) and Root Mean Squared Error ($MSE$) are chosen to quantify the counting performance. They are defined as
	% \begin{equation}\label{MAE}
	% \small
	% MAE=\frac{1}{N} \sum_{i=1}^{N} |C^{pre}_{i}-C^{gt}_{i}|\,,
	% \end{equation}
	% \begin{equation}\label{RMSE}
	% \small
	% MSE=\sqrt{ \frac{1}{N} \sum_{i=1}^{N} (C^{pre}_{i}-C^{gt}_{i})^{2}}\,,
	% \end{equation}
	% where $N$ denotes the number of images, $C^{pre}_{i}$ denotes the predicted count of the $i$-th image, and $C^{gt}_{i}$ denotes the corresponding ground-truth count. $MAE$ measures the accuracy of counting, and $MSE$ measures the stability. Lower $MAE$ and $MSE$ imply better counting performance.
	
	\subsection{Implementation Details}
	
	%\vspace{-10pt}
	\paragraph{Interval Partition}\label{One_two_linear}
	
	We generate ground-truth counts of local patches by integrating over the density maps. The counts are usually not integers, because objects can partly present in cropped local patches. %it is necessary to count part of the objects as $0.5$ at least. 
	We evaluate two different partition strategies. In the first partition, we choose $0.5$ as the step and generate partitions as $\{0\}$, $(0.0.5]$,$(0.5,1]$, ... ,$(C_{max}-0.5,C_{max}]$ and $(C_{max},+\infty)$, where $C_{max}$ denotes the maximum count of the closed set. This partition is named as \texttt{One-Linear Partition}.
	
	In the second partition, we further finely divide the sub-interval $(0.0.5]$, because this interval contains a sudden change from no object to part of an object, and a large proportion of objects lie in this sub-interval. A small step of $0.05$ is used to divide this interval. We call this partition \texttt{Two-Linear Partition}.  
	
	% Both partitions will be evaluated in the following experiments.

	\vspace{-10pt}
	\paragraph{Data Augmentation}
	We follow the same data augmentation used in~\cite{CSRNet_2018_CVPR}, except for the UCF-QNRF dataset~\cite{Compose_Loss_2018_ECCV}. In particular, $9$ sub-images of $\frac{1}{4}$ resolution are cropped from the original image. The first $4$ sub-images are from four corners, and the remaining $5$ are randomly cropped. Random scaling and mirroring are also performed. For the UCF-QNRF dataset~\cite{Compose_Loss_2018_ECCV}, we follow the same setting as in~\cite{Compose_Loss_2018_ECCV} and crop the original image into $224\times224$ sub-images.
	
	\vspace{-10pt}
	\paragraph{Training Details}
	S-DCNet is implemented with \texttt{PyTorch}. We train S-DCNet using standard SGD. The encoder in S-DCNet is directly adopted from convolutional layers of VGG16~\cite{Simonyan2014Very_VGG16} pretrained on ImageNet, and the other layers employ random Gaussian initialization with a standard deviation of $0.01$. The learning rate is initially set to $0.001$ and is decreased by $\times10$ when the training error stagnates. We keep training until convergence. For the ShanghaiTech, UCF\_CC\_50, TRANCOS and MTC datasets, the batch size is set to $1$. For the UCF-QNRF dataset, the batch size is set to 16 following~\cite{Compose_Loss_2018_ECCV}. 
	
	\subsection{Ablation Study on the ShanghaiTech Part\_A}
	%\vspace{-10pt}
	%\paragraph{$(1)$ Is S-DCNet robust to $C_{max}$}
	\paragraph{Is S-DCNet Robust to $C_{max}$?}
	%here discuss about regress and classification, and two kind of interval divide
	
	%1.compare different threshold setting for classification, only divide and counting can be robust (3 times: 5,4,3)
	When reformulating the counting problem into classification, a critical issue is how to choose $C_{max}$, which defines the closed set. Hence, it is important that S-DCNet is robust to the choice of $C_{max}$.
	
	We conduct a statistical analysis on count values of local patches in the training set, and then set $C_{max}$ with the quantiles ranging from $100\%$ to $80\%$ (decreased by $5\%$). Two-stage S-DCNet is evaluated. Another baseline of classification without S-DC is also used to explore whether counting can be simply modeled in a closed-set classification manner. To be specific, we reserve the VGG16 encoder and the classifier in this classification baseline.
	
\begin{figure}[t]
		\begin{center}
			\includegraphics[width=0.8\linewidth]{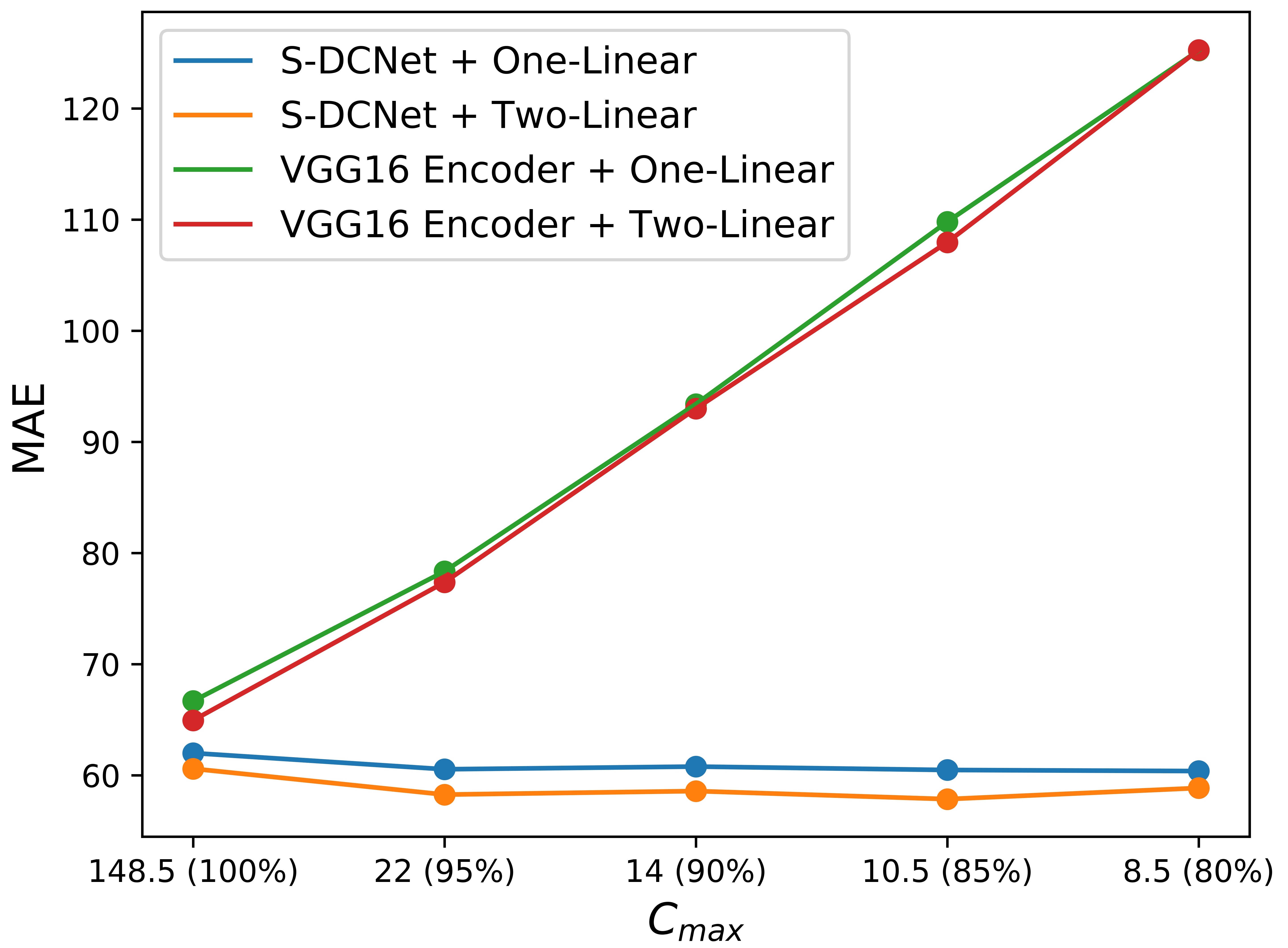}
		\end{center}
		\vspace{-10pt}
		\caption{The influence of $C_{max}$ to S-DCNet on the ShanghaiTech Part\_A dataset~\cite{MCNN_2016_CVPR}. 
			% 	$C_{max}$ represents the maximum count of the classifier. 
			The numbers in the brackets denote quantiles of the training set, for example, $22$ ($95\%$) means the $95\%$ quantile is $22$. `VGG16~Encoder' is the classification baseline without \mbox{S-DC}. 
			%It sets up the baseline of classification without S-DC for S-DCNet. 
			`One-Linear' and `Two-Linear' are defined in Section~\ref{One_two_linear}.}
		\label{fig:mae_versus_cmax}
		\vspace{-10pt}
	\end{figure}

	Results are presented in Figure~\ref{fig:mae_versus_cmax}. It can be observed that the MAE of the classification baseline increases rapidly with decreased $C_{max}$. This result is not surprising, because the model is constrained to be visible to count values not greater than $C_{max}$.  This suggests that counting cannot be simply transformed into closed-set classification. However, with the help of S-DC, S-DCNet exhibits strong robustness to the changes of $C_{max}$. It seems the systematic error brought by $C_{max}$ can somewhat be alleviated with S-DC. 
	Regarding how to choose concrete $C_{max}$, the maximum count of the training set seems not the best choice, while some small quantiles even deliver better performance. Perhaps a model is only able to count objects accurately within a certain degree of denseness. %Maybe it is easier to count larger numbers through S-DC rather than directly recognizing larger counts. 
	We also notice \texttt{Two-Linear Partition} is slightly better than \texttt{One-Linear Partition}, which indicates that the fine division to the $(0,0.5]$ sub-interval has a positive effect.
	% justifies the opinion in Section~\ref{One_two_linear}. 
	
	\begin{figure*}
		\begin{minipage}{\textwidth}
			
			%SHAB
			\makeatletter\def\@captype{table}\makeatother
			\begin{minipage}{.23\textwidth}
				\centering
				\footnotesize
				\begin{tabular}{|c|c|c|}
					\hline
					Division time & MAE &MSE\\
					\hline
					0 & 76.0 & 142.5\\		  
					1 & 62.2 & 103.4\\
					2 & \textbf{58.3} & \textbf{95.0}\\
					3 & 60.1 & 99.8\\	 
					4 & 61.9 & 107.2\\
					\hline
				\end{tabular}
				\vspace{1pt}
				\caption{Results of S-DCNet with different S-DC stages. The best performance is boldfaced.\newline}
				\label{tab:div_time}
				% 	\vspace{-5pt}
			\end{minipage}
			\hfill
			%UCFCC50
			\makeatletter\def\@captype{table}\makeatother
			\begin{minipage}{.34\textwidth}
				\centering
				\footnotesize
				\begin{tabular}{|l|c|c|c|c|c|c|c|}
					\hline
					Method           & MAE &MSE\\
					\hline
					
					classification   & 77.4 &  149.3  \\   
					regression       & 68.9 & 112.1   \\
					open-set regression + S-DC & 66.6 & 107.9	   \\  
					closed-set regression + S-DC  &64.7  & 105.7   \\ \hline
					S-DCNet (2)      & \textbf{58.3} & \textbf{95.0}\\
					
					\hline
				\end{tabular}
				\vspace{5pt}
				\caption{Effect of S-DC. Two classification and regression baselines are compared against S-DCNet. S-DCNet (2) denotes two-stage S-DCNet. The best performance is boldfaced.}
				% 	\vspace{-5pt}
				\label{tab:reg_cls_div} 
			\end{minipage}
			\hfill
			%UCF-QNRF
			\makeatletter\def\@captype{table}\makeatother
			\begin{minipage}{.34\textwidth}
				\centering
				\footnotesize
				\addtolength{\tabcolsep}{4.5pt}
				\begin{tabular}{|c|c|c|c|}
					%\hline
					%\multicolumn{2}{|c|}{Loss} & &\\
					\hline
					$\sum_{i=0}^2L_C^i$  &$L_R^2$   & MAE   &MSE\\
					\hline
					&\checkmark& 301.4& 396.9  \\
					\checkmark&          &88.4	 & 128.8   \\
					\checkmark &\checkmark& \textbf{58.3} & \textbf{95.0}\\
					
					\hline
				\end{tabular}
				\vspace{5pt}
				\caption{Effect of different loss functions. Note that, multi-stage predictions are averaged if $L_R^2$ is not applied, because the division decider cannot receive supervision signal during training. The best performance is boldfaced.\newline}
				% 	\vspace{-5pt}
				\label{tab:loss_ablation} 
			\end{minipage}
			
		\end{minipage}
		\vspace{-10pt}
	\end{figure*}
	
	According to the above results, S-DCNet is robust to $C_{max}$ in a wide range of values, and $C_{max}$ is generally encouraged to be set less than the maximum count value observed. In addition, there is no significant difference between two kinds of partitions. For simplicity, we set $C_{max}$ to be the $95\%$ quantile and adopt \texttt{Two-Linear Partition} in the following experiments.
	
	\vspace{-10pt}
	\paragraph{How Many Times to Divide?}
	%how many times to divide is better (4 or 5 times)
	S-DCNet can apply S-DC up to $4$ times, but how many times are sufficient? Here we evaluate S-DCNet with different division stages. Quantitative results are listed in Table~\ref{tab:div_time}. It can be observed that applying two-stage S-DC is clearly adequate.
	
	\vspace{-10pt}
	\paragraph{The Effect of S-DC}
	To highlight the effect of S-DC, we compare S-DCNet against several regression and classification baselines. These baselines adopt the same architecture of VGG16 encoder and the classifier in S-DCNet. \textit{classification} is the result of $C_0$ without S-DC, and $C_{max}$ is set to be the $95\%$ quantile ($22$). For all regression baselines, we modify the output channel of the classifier to be $1$ and employ the $\ell_1$ loss. We set three regression baselines. \textit{regression} predicts counts without S-DC. To justify whether S-DC can also work in regression, we adapt the S-DC idea to regression under both open-set and closed-set settings. \textit{open-set regression + S-DC} is straight-forward. We do not limit the output range, and it can vary from $0$ to $+\infty$. \textit{closed-set regression + S-DC} indicates that the output range is constrained within $[0,C_{max}]$ ($C_{max}$ is set to $22$ for a fair comparison), and large outputs will be clipped to $C_{max}$.
	
	Results are shown in Table~\ref{tab:reg_cls_div}. We can see that counting by classification without S-DC suffers from the limitation of $C_{max}$ and performs even worse than regression. In addition, regression can also benefit from S-DC, and it is encouraged to limit the output range of the regressor in a closed set. Moreover, with S-DC, S-DCNet significantly reduces the counting error and outperforms both the classification and regression baselines by a large margin. This verifies our argument that it is more effective to reformulate counting in classification than in regression. Perhaps the optimization is easier and less sensitive to sample imbalance in classification than in regression. Whatever, at least one thing is made clear: a counting model can learn from a closed set and generalize well to a open set via S-DC.

\begin{figure}[t]
		\begin{center}
			\includegraphics[width=0.8\linewidth]{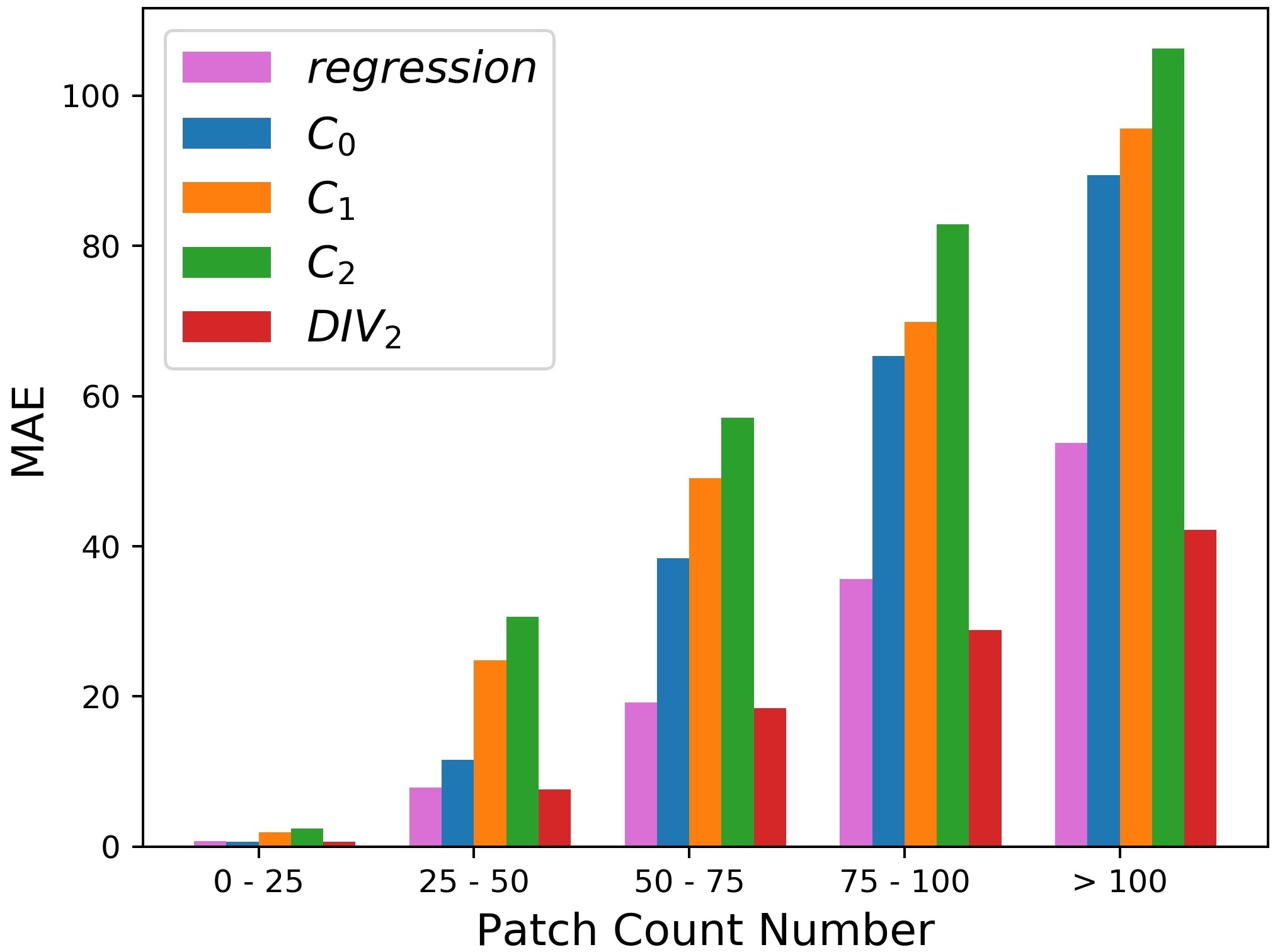}
		\end{center}
		\vspace{-10pt}
		\caption{Counting errors of $64\times64$ local patches on the test set of ShanghaiTech Part\_A~\cite{MCNN_2016_CVPR}. $regression$ denotes direct local counts regression using VGG16. $C_0$, $C_1$ and $C_2$ are single-branch predictions conditioned on $F_0$, $F_1$ and $F_2$, respectively. $DIV_2$ denotes two-stage S-DCNet, which fuses the predictions of $C_0$, $C_1$ and $C_2$ with S-DC.}
		\label{fig:patch_div_err}
		\vspace{-10pt}
	\end{figure}

	We further analyze the counting error of $64\times64$ local patches in detail. As shown in Figure~\ref{fig:patch_div_err}, we observe that the direct single-branch prediction without S-DC (predicting $C_0$, $C_1$ and $C_2$ from $F0$, $F1$ and $F2$, respectively) performs worse than the regression baseline, which can be attributed to the limited $C_{max}$ of the classifier. After embedding the S-DC strategy to divide and merge the count map of multiple resolutions, counting errors significantly reduce. Such a benefit is even much obvious in dense patches with local counts greater than $100$. It justifies our argument that, instead of regressing a large count value directly, it is more accurate to count dense patches through S-DC. 
	
	%2.represent the counting error of the patch (for differnt number level) between simply classification and divide and conquer (3 times: 5,4,3)
	
	% \begin{table}\footnotesize
	% 	\begin{center}
	% 	\footnotesize
	% 		\begin{tabular}{|c|c|c|c|}
	% 			%\hline
	% 			%\multicolumn{2}{|c|}{Loss} & &\\
	% 			\hline
	% 			 $\sum_{i=0}^2L_C^i$  &$L_R^2$   & MAE   &MSE\\
	% 			\hline
	% 			           &\checkmark& 301.4& 396.9  \\
	% 			 \checkmark&          &88.4	 & 128.8   \\
	%              \checkmark &\checkmark& \textbf{58.3} & \textbf{95.0}\\
	
	% 			\hline
	% 		\end{tabular}
	% 	\end{center}
	% 	\caption{Effect of different loss functions. Note that, multi-stage predictions are averaged if $L_R^2$ is not applied, because the division decider cannot receive supervision signal during  %training. The best performance is boldfaced.}
	% 	\vspace{-10pt}
	% 	\label{tab:loss_ablation}
	% \end{table}   
	
	\vspace{-10pt}
	\paragraph{Loss Functions Also Matter}
	We further validate the effect of different loss functions used in S-DCNet and report the results in Table~\ref{tab:loss_ablation}. S-DCNet works poorly when trained with only $L_R^2$. This is not surprising, because no supervision signal is provided to multi-stage division results. In addition, it seems necessary for the division decider to decide where to divide, because S-DCNet greatly benefits from the help of merging loss $L_R^2$.
	
	Through the visualizations of $W_i$s in Fig.~\ref{fig:w_visual}, we observe that reasonably good divisions can be achieved with the supervision of $L_R^2$. This has another benefit, the network can learn when to divide not just in counts larger than $C_{max}$.
	
	\begin{figure}[!t]
		\begin{center}
			%\fbox{\rule{0pt}{1.8in} \rule{0.9\linewidth}{0pt}}
			\includegraphics[width=1.0\linewidth]{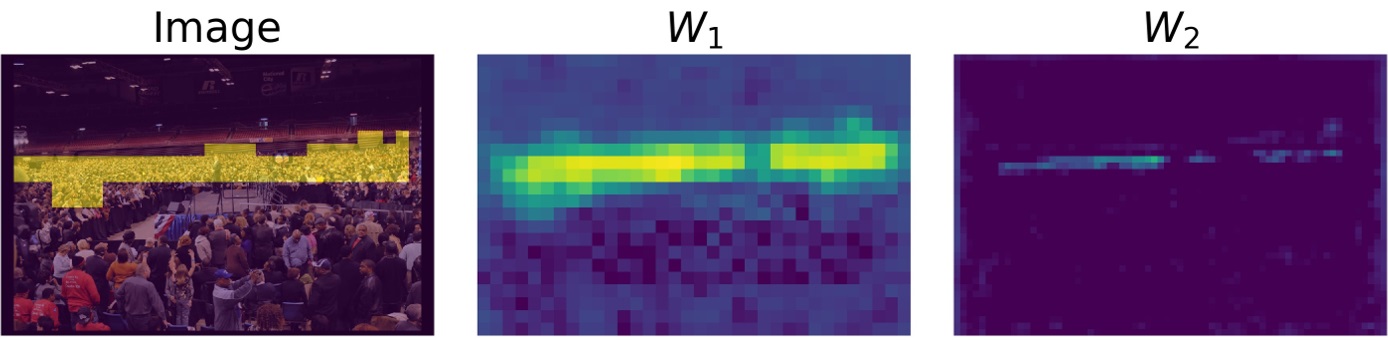}
		\end{center}
		\vspace{-10pt}
		\caption{Visualizations of $W_i$s in S-DCNet. The brighter the image is, the greater the values are. In the input image, count values greater than $C_{max}$ are indicated by yellow regions. It is clear that $W_i$ appropriately identifies regions to be divided.}
		\label{fig:w_visual}
	\end{figure}

	%\subsubsection{Visualization of feature maps with different resolutions}
	%\subsubsection{How does division work?}

	\subsection{Comparison with State of the Art}
	%first use a tab to point the best setting for Spatial DCNet, including interval partition, $C_{max}$ setting and division times.
	According to the ablation study, the final configurations for S-DCNet are summarized in Table~\ref{tab:compare_setting}. Qualitative results are shown in the Supplement.

	\begin{table}\footnotesize
		\centering
		\addtolength{\tabcolsep}{4pt}
		\begin{tabular}{|l|c|c|c|c|}
			\hline
			Dataset & $C_{max}$ &max& Gaussian kernel\\
			\hline
			SH Part\_A~\cite{MCNN_2016_CVPR}&22.0	& 148.5 &\multirow{3}{*}{Geometry-Adaptive}\\ \cline{1-3} 
			UCF\_CC\_50~\cite{UCFCC50_2013_CVPR}& $-$ & $-$ &\\	\cline{1-3} 
			UCF-QNRF~\cite{Compose_Loss_2018_ECCV}& 8.0 & 131.5& \\ \cline{1-4}
			SH Part\_B~\cite{MCNN_2016_CVPR}& 7.0 & 83.0&  Fixed: $\sigma=15$\\ \cline{1-4}
			Trancos~\cite{TRANCOSdataset_IbPRIA2015}& 5.0 & 24.5& Fixed: $\sigma=10$\\	   \cline{1-4}
			MTC~\cite{Lu2017TasselNet}& 3.5 & 8.0& Fixed: $\sigma=8$\\	   \cline{1-4}
			\hline
			Partition& \multicolumn{3}{c|}{Two-Linear}\\ \cline{1-4}
			Type of~$C_{max}$ &\multicolumn{3}{c|}{$95\%$ quantile}\\ \cline{1-4}
			\hline
		\end{tabular}
		\vspace{5pt}
		\caption{Overall configurations of S-DCNet. $max$ denotes the maximum count of local patch in the training set, while $C_{max}$ is the maximum count set for the closed set in S-DCNet. $Gaussian~kernel$ is used to generate density maps from dotted annotations. Specially, since UCF\_CC\_50 adopts 5-fold cross-validation, $max$ and $C_{max}$ are set differently for each fold.}
		\label{tab:compare_setting}
		\vspace{-10pt}
	\end{table}

	\begin{figure*}
		\begin{minipage}{\textwidth}
			
			%SHAB
			\makeatletter\def\@captype{table}\makeatother
			\begin{minipage}{.38\textwidth}
				\centering
				\footnotesize
				\begin{tabular}{|l|c|c|c|c|}
					\hline
					&\multicolumn{2}{c|}{Part A} &
					\multicolumn{2}{c|}{Part B}\\
					\hline
					Method & MAE &MSE& MAE &MSE\\
					\hline
					Zhang \textit{et~al.}~\cite{Zhang_2015_CVPR}&181.8&277.7&32.0&49.8\\
					%MCNN~\cite{}&110.2&173.2&26.4&41.3\\
					%Switching-CNN~\cite{}&90.4&135.0&21.6&33.4\\
					CP-CNN~\cite{CPCNN_2017_ICCV} &73.6 & 106.4 &20.1&30.1\\
					D-ConvNet~\cite{DeepNegCor_2018_CVPR} &73.5	&112.3&18.7	&26.0\\
					IG-CNN~\cite{Divide_grow_2018_CVPR}&72.5	&118.2	&13.6&21.1\\
					DRSAN~\cite{DRSAN2018Crowd}&69.3	&96.4 &11.1&18.2\\
					CSRNet~\cite{CSRNet_2018_CVPR}&68.2 & 115.0 &10.6&16.0\\
					SANet~\cite{SANet_2018_ECCV}&67.0	&104.5 &8.4&13.6\\
					SPN~\cite{SPN_2019_WACV}&61.7&99.5&9.4&14.4\\
					%ASD	~\cite{}&65.6	&98.0 &8.5 &13.7\\		
					\hline
					S-DCNet & \textbf{58.3} & \textbf{95.0}& \textbf{6.7}	  & \textbf{10.7}  \\	   
					\hline
				\end{tabular}  
				\vspace{5pt}
				\caption{Comparison with state-of-the-art approaches on the test set of ShanghaiTech~\cite{MCNN_2016_CVPR} dataset. The best performance is boldfaced.}
				\label{tab:compare_SHAB}
			\end{minipage}
			\hfill
			%UCFCC50
			\makeatletter\def\@captype{table}\makeatother
			\begin{minipage}{.25\textwidth}
				\centering
				\footnotesize
				\begin{tabular}{|l|c|c|}
					\hline
					Method & MAE &MSE\\
					\hline
					Idrees\textit{et~al.}~\cite{UCFCC50_2013_CVPR}&468.0&590.3\\
					Zhang\textit{et~al.}~\cite{Zhang_2015_CVPR}&467.0&498.5\\
					IG-CNN~\cite{Divide_grow_2018_CVPR}&291.4&349.4\\
					D-ConvNet~\cite{DeepNegCor_2018_CVPR}&288.4&404.7\\
					CSRNet~\cite{CSRNet_2018_CVPR}&266.1& 397.5\\
					
					SANet~\cite{SANet_2018_ECCV}&258.4	&334.9\\
					DRSAN~\cite{DRSAN2018Crowd}&219.2	&\textbf{250.2}\\
					%ASD	~\cite{}&\textbf{196.2}	&270.9 \\	
					\hline
					S-DCNet & \textbf{204.2} & 301.3\\
					
					\hline
				\end{tabular}
				\vspace{0pt}
				\caption{Comparison with state-of-the-art approaches on the test set of UCF\_CC\_50~\cite{UCFCC50_2013_CVPR} dataset. The best performance is boldfaced.\newline}
				\label{tab:compare_UCF_CC}    
			\end{minipage}
			\hfill
			%UCF-QNRF
			\makeatletter\def\@captype{table}\makeatother
			\begin{minipage}{.30\textwidth}
				\centering
				\footnotesize
				\begin{tabular}{|l|c|c|}
					\hline
					Method & MAE &MSE\\
					\hline
					Idrees\textit{et~al.}~\cite{UCFCC50_2013_CVPR}&315&508\\
					MCNN~\cite{MCNN_2016_CVPR}&277&426\\
					Encoder-Decoder~\cite{En_De2017segnet}& 270 &478\\
					CMTL~\cite{CMTL2017CNN}&252&514\\
					Switching-CNN~\cite{SwitchCNN_2017_CVPR}&228&445\\
					Base Network~\cite{Compose_Loss_2018_ECCV}& 163& 227\\
					Composition Loss~\cite{Compose_Loss_2018_ECCV}&132&191\\
					\hline
					S-DCNet & \textbf{104.4}&\textbf{176.1}	\\
					
					\hline
				\end{tabular}
				\vspace{5pt}
				\caption{Comparison with state-of-the-art approaches on the test set of UCF-QNRF~\cite{Compose_Loss_2018_ECCV} dataset. The best performance is boldfaced.\newline}
				\label{tab:compare_UCF-QNRF}    
			\end{minipage}
			\vfill
			\vspace{5pt}
			%TRANCOS
			\makeatletter\def\@captype{table}\makeatother
			\begin{minipage}{.5\textwidth}
				\centering
				\footnotesize
				\addtolength{\tabcolsep}{2pt}
				\begin{tabular}{|l|c|c|c|c|}
					\hline
					Method &GAME(0) &GAME(1)&GAME(2)&GAME(3)\\
					\hline
					%Lempitsky \etal~\cite{vlaz2010denlearn}&13.76 &16.72 &20.72 &24.36\\
					CCNN~\cite{O2016Towards_CCNN}&12.49 &16.58 &20.02 &22.41\\
					Hydra-3s~\cite{O2016Towards_CCNN}&10.99 &13.75 &16.69 &19.32\\
					CSRNet~\cite{CSRNet_2018_CVPR}&3.56 &5.49 &8.57 &15.04\\
					SPN~\cite{SPN_2019_WACV}&3.35 &4.94 &6.47 &9.22\\
					\hline
					S-DCNet & \textbf{2.92} &\textbf{4.29} &\textbf{5.54} &\textbf{7.05}\\
					\hline
				\end{tabular}
				\vspace{0pt}
				\caption{Comparison with state-of-the-art approaches on the test set of TRANCOS~\cite{TRANCOSdataset_IbPRIA2015} dataset. The best performance is boldfaced.}
				\label{tab:compare_Trancos}    
			\end{minipage}
			\hfill
			%MTC
			\makeatletter\def\@captype{table}\makeatother
			\begin{minipage}{.46\textwidth}
				\centering
				\footnotesize
				\addtolength{\tabcolsep}{2pt}
				\begin{tabular}{|l|c|c|}
					\hline
					Method & MAE &MSE\\
					\hline
					
					%JointSeg~\cite{lu2016region}	           & 24.2 & 31.6\\
					%mTASSEL~\cite{lu2015fine}               & 19.6 & 26.1\\
					GlobalReg~\cite{tota2015counting}             & 19.7 & 23.3\\
					DensityReg~\cite{vlaz2010denlearn}             & 11.9 & 14.8\\
					CCNN~\cite{O2016Towards_CCNN}                  & 21.0 & 25.5\\
					TasselNet~\cite{Lu2017TasselNet}		       & 6.6 & 9.6\\
					\hline
					S-DCNet & \textbf{5.6} &\textbf{9.1}\\
					\hline
				\end{tabular}
				\vspace{5pt}
				\caption{Comparison with state-of-the-art approaches on the test set of MTC~\cite{Lu2017TasselNet} dataset. The best performance is boldfaced.}
				\label{tab:compare_MTC}    
			\end{minipage}
			\hfill

		\end{minipage}
		\vspace{-10pt}
	\end{figure*}

	\vspace{-10pt}
	\paragraph{The ShanghaiTech Dataset}
	The ShanghaiTech crowd counting dataset~\cite{MCNN_2016_CVPR} is consisted of two parts: Part\_A and Part\_B. Part\_A includes $300$ images for training and $182$ for testing. This part represents highly congested scenes. Part\_B contains $716$ images in relatively sparse scenes, where $400$ images are used for training and $316$ for testing. Quantitative results are listed in Table~\ref{tab:compare_SHAB}. Our method outperforms the previous state-of-the-art SPN~\cite{SPN_2019_WACV} and SANet~\cite{SANet_2018_ECCV} with a $5.5\%$ relative improvement in Part\_A and $20.2\%$ in Part\_B, respectively. These results suggest S-DCNet is able to adapt to both sparse and crowded scenes. 
	
	%--------------------------------------------------------------------------------------------
	\iffalse
	\begin{table}\footnotesize
		\begin{center}
			\begin{tabular}{|l|c|c|c|c|}
				\hline
				&\multicolumn{2}{c|}{Part A} &
				\multicolumn{2}{c|}{Part B}\\
				\hline
				Method & MAE &MSE& MAE &MSE\\
				\hline\hline
				Zhang \textit{et~al.}~\cite{Zhang_2015_CVPR}&181.8&277.7&32.0&49.8\\
				%MCNN~\cite{}&110.2&173.2&26.4&41.3\\
				%Switching-CNN~\cite{}&90.4&135.0&21.6&33.4\\
				CP-CNN~\cite{CPCNN_2017_ICCV} &73.6 & 106.4 &20.1&30.1\\
				D-ConvNet~\cite{DeepNegCor_2018_CVPR} &73.5	&112.3&18.7	&26.0\\
				IG-CNN~\cite{Divide_grow_2018_CVPR}&72.5	&118.2	&13.6&21.1\\
				DRSAN~\cite{DRSAN2018Crowd}&69.3	&96.4 &11.1&18.2\\
				CSRNet~\cite{CSRNet_2018_CVPR}&68.2 & 115.0 &10.6&16.0\\
				SANet~\cite{SANet_2018_ECCV}&67.0	&104.5 &8.4&13.6\\
				SPN~\cite{SPN_2019_WACV}&61.7&99.5&9.4&14.4\\
				%ASD	~\cite{}&65.6	&98.0 &8.5 &13.7\\			
				S-DCNet (ours) & \textbf{58.3} & \textbf{95.0}& \textbf{6.7}	  & \textbf{10.7}  \\	   
				\hline
			\end{tabular}
		\end{center}
		\caption{Comparison with State-of-the-Art Counting Approaches on the Test Set of ShanghaiTech~\cite{MCNN_2016_CVPR} dataset. The best performance is boldfaced.}
		\label{tab:compare_SHAB}
	\end{table}  
	\fi
	%--------------------------------------------------------------------------------------------
	
	\vspace{-10pt}
	\paragraph{The UCF\_CC\_50 Dataset}
	
	UCF\_CC\_50~\cite{UCFCC50_2013_CVPR} is a tiny crowd counting dataset with $50$ images in extremely crowded scenes. The number of people within an images varies from $96$ to $4633$. We follow the 5-fold cross-validation as in~\cite{UCFCC50_2013_CVPR}. Results are shown in Table~\ref{tab:compare_UCF_CC}. Our method surpasses the previous best method, DRSAN~\cite{DRSAN2018Crowd}, with a $6.8\%$ relative improvement in \textit{MAE}.%Our method is comparable with current optimal ASD~\cite{} while greatly surpass the other methods.  
	
	%--------------------------------------------------------------------------------------------
	\iffalse
	\begin{table}\footnotesize
		\begin{center}
			\begin{tabular}{|l|c|c|}
				\hline
				Method & MAE &MSE\\
				\hline\hline
				Idrees\textit{et~al.}~\cite{UCFCC50_2013_CVPR}&468.0&590.3\\
				Zhang\textit{et~al.}~\cite{Zhang_2015_CVPR}&467.0&498.5\\
				IG-CNN~\cite{Divide_grow_2018_CVPR}&291.4&349.4\\
				D-ConvNet~\cite{DeepNegCor_2018_CVPR}&288.4&404.7\\
				CSRNet~\cite{CSRNet_2018_CVPR}&266.1& 397.5\\
				
				SANet~\cite{SANet_2018_ECCV}&258.4	&334.9\\
				DRSAN~\cite{DRSAN2018Crowd}&219.2	&\textbf{250.2}\\
				%ASD	~\cite{}&\textbf{196.2}	&270.9 \\	
				
				S-DCNet (ours) & \textbf{204.2} & 301.3\\
				
				\hline
			\end{tabular}
		\end{center}
		\caption{Comparison with State-of-the-Art Counting Approaches on the Test Set of UCF\_CC\_50~\cite{UCFCC50_2013_CVPR} dataset. The best performance is boldfaced.}
		\label{tab:compare_UCF_CC}
	\end{table} 
	\fi
	%--------------------------------------------------------------------------------------------
	
	\vspace{-10pt}
	\paragraph{The UCF-QNRF Dataset} 
	UCF-QNRF~\cite{Compose_Loss_2018_ECCV} is a large crowd counting dataset with $1535$ high-resolution images and $1.25$ million head annotations. There are $1201$ training images and $334$ test images. It contains extremely congested scenes where the maximum count of an image can reach $12865$. We follow the same image processing as in~\cite{Compose_Loss_2018_ECCV} and report results in Table~\ref{tab:compare_UCF-QNRF}. Our method reaches the state-of-the-art performance and surpasses the previous best method with a $20.9\%$ boost in \textit{MAE}. We surprisingly notice that S-DCNet only learn from a closed set with $C_{max}=8.0$, which is only $6\%$ of the maximum count $131.5$ according to Table~\ref{tab:compare_setting}. S-DCNet, however, generalizes to large counts effectively and predicts accurate counts. 
	% \emph{Maybe it is better to adopt spatial divide-and-conquer strategy to transfer counting from open-set regression to closed-set classification.}     
	
	%--------------------------------------------------------------------------------------------
	\iffalse
	\begin{table}\footnotesize
		\begin{center}
			\begin{tabular}{|l|c|c|}
				\hline
				Method & MAE &MSE\\
				\hline\hline 
				Idrees\textit{et~al.}~\cite{UCFCC50_2013_CVPR}&315&508\\
				MCNN~\cite{MCNN_2016_CVPR}&277&426\\
				Encoder-Decoder~\cite{En_De2017segnet}& 270 &478\\
				CMTL~\cite{CMTL2017CNN}&252&514\\
				Switching-CNN~\cite{SwitchCNN_2017_CVPR}&228&445\\
				Base Network~\cite{Compose_Loss_2018_ECCV}& 163& 227\\
				Composition Loss~\cite{Compose_Loss_2018_ECCV}&132&191\\
				S-DCNet (ours) & \textbf{104.4}&\textbf{176.1}	\\
				
				\hline
			\end{tabular}
		\end{center}
		\caption{Comparison with State-of-the-Art Counting Approaches on the Test Set of UCF-QNRF~\cite{Compose_Loss_2018_ECCV} dataset. The best performance is boldfaced.}
		\label{tab:compare_UCF-QNRF}
	\end{table}  
	\fi
	%--------------------------------------------------------------------------------------------
	
	\vspace{-10pt}
	\paragraph{The TRANCOS Dataset}
	Aside from crowd counting, we also evaluate S-DCNet on a vehicle counting dataset, TRANCOS~\cite{TRANCOSdataset_IbPRIA2015}, to see its generalization ability. TRANCOS contains $1244$ images of congested traffic scenes in various perspectives. It adopts the Grid Average Mean Absolute Error (GAME)~\cite{TRANCOSdataset_IbPRIA2015} as the evaluation metric. $GAME(L)$ divides an image into $2^L\times2^L$ non-overlapping sub-regions and accumulates of the $MAE$ over sub-regions. Larger $L$ implies better local predictions. In particular, $GAME(0)$ downgrades to $MAE$. 
	%which emphasizes the accuracy of local regions.
	\iffalse
	which is defined as
	\begin{equation}\label{Game_loss}
	GAME(L) = \frac{1}{N}\sum\limits_{n=1}^{N}(\sum\limits_{l=1}^{4^L}|C_{pre}^l-C_{gt}^l|)\,,
	\end{equation}
	where $N$ denotes the number of images. $C_{pre}^l$ and $C_{gt}^l$ are the predicted and ground-truth count of the $L$-th sub-region, respectively. 
	\fi
	Results are listed in Table~\ref{tab:compare_Trancos}. S-DCNet surpasses other methods under all $GAME(L)$ metrics, and particularly, delivers a $22.5\%$ relative improvement on $GAME(3)$. This suggests S-DCNet not only achieves accurate global predictions but also behaves well in local regions. 
	
	%--------------------------------------------------------------------------------------------
	\iffalse
	\begin{table}\footnotesize
		\begin{center}
			
			\begin{tabular}{|l|c|c|c|c|}
				\hline
				Method &GAME(0) &GAME(1)&GAME(2)&GAME(3)\\
				\hline\hline
				%Lempitsky \etal~\cite{vlaz2010denlearn}&13.76 &16.72 &20.72 &24.36\\
				CCNN~\cite{O2016Towards_CCNN}&12.49 &16.58 &20.02 &22.41\\
				Hydra-3s~\cite{O2016Towards_CCNN}&10.99 &13.75 &16.69 &19.32\\
				CSRNet~\cite{CSRNet_2018_CVPR}&3.56 &5.49 &8.57 &15.04\\
				SPN~\cite{SPN_2019_WACV}&3.35 &4.94 &6.47 &9.22\\
				S-DCNet (ours) & \textbf{2.92} &\textbf{4.29} &\textbf{5.54} &\textbf{7.05}\\
				
				\hline
			\end{tabular}
		\end{center}
		\caption{Comparison with State-of-the-Art Counting Approaches on the Test Set of TRANCOS~\cite{TRANCOSdataset_IbPRIA2015} dataset. The best performance is boldfaced.}
		\label{tab:compare_Trancos}
	\end{table}
	\fi
	%--------------------------------------------------------------------------------------------
	
	\vspace{-10pt}
	\paragraph{The MTC Dataset}
	We further evaluate our method on a plant counting dataset, i.e., the MTC dataset~\cite{Lu2017TasselNet}. The MTC dataset contains $361$ high-resolution images of maize tassels collected from 2010 to 2015 in the wild field. In contrast to people or vehicles that have similar physical sizes, maize tassels are with heterogeneous physical sizes and are self-changing over time. We think this dataset is suitable for justifying the robustness of S-DCNet to object-size variations. We follow the same setting as in~\cite{Lu2017TasselNet} and report quantitative results in Table~\ref{tab:compare_MTC}. Although the previous best method, TasselNet~\cite{Lu2017TasselNet}, already exhibits accurate results, S-DCNet still shows a certain degree of improvement. 
	
	%--------------------------------------------------------------------------------------------
	\iffalse\begin{table}\footnotesize
		\begin{center}
			\begin{tabular}{|l|c|c|}
				\hline
				Method & MAE &MSE\\
				\hline\hline
				
				%JointSeg~\cite{lu2016region}	           & 24.2 & 31.6\\
				%mTASSEL~\cite{lu2015fine}               & 19.6 & 26.1\\
				GlobalReg~\cite{tota2015counting}             & 19.7 & 23.3\\
				DensityReg~\cite{vlaz2010denlearn}             & 11.9 & 14.8\\
				CCNN~\cite{O2016Towards_CCNN}                  & 21.0 & 25.5\\
				TasselNet~\cite{Lu2017TasselNet}		       & 6.6 & 9.6\\
				S-DCNet (ours) & \textbf{5.6} &\textbf{9.1}\\
				
				\hline
			\end{tabular}
		\end{center}
		\caption{Comparison with State-of-the-Art Counting Approaches on the Test Set of MTC~\cite{Lu2017TasselNet} dataset. The best performance is boldfaced.}
		\label{tab:compare_MTC}
	\end{table}
	\fi
	%--------------------------------------------------------------------------------------------
	
	%\subsubsection{Trancos dataset}
	
	\section{Conclusion}
	Counting is an open-set problem in theory, but only a finite closed set can be observed in reality. This is particularly true because any dataset is always a sampling of the real world. Inspired by the decomposable property of counting, we propose to transform the open-set counting into a closed-set problem, and address the problem with the idea of S-DC. We realize S-DC in a deep counting network termed S-DCNet. We show through a toy experiment and extensive evaluations on standard benchmarks that, even given a closed training set, S-DCNet can effectively generalize to open-set scenarios.
	
	For future work, we will test the adaptability of S-DC on other network architectures.
	%The count number of objects in real life is an open set but only a finite closed set of patterns can be observed. It is challenging for counting networks to adpat to the open set while only learn %from a closed set. We proposed a simple but effective S-DCNet to alleviate such a challenge. S-DCNet learns to count a closed set and generalize to the open set via spatial divide-and-conquer (S-DC). %Extensive experiments demonstrate significant improvements brought by our method. What's more, the S-DC strategy can be easily embedded in other counting approaches. 
	
	\vspace{-10pt}
	\paragraph{Acknowledgements}
	This work was supported by the Natural Science Foundation of China under Grant No. 61876211.

	{\small
		\bibliographystyle{ieee_fullname}
		\bibliography{egbib}
	}

\section{Supplementary Materials}
	In this Supplement, we provide further clarifications and discussions on the motivation of S-DCNet, compare S-DCNet with other related ideas, and show qualitative results on evaluated datasets.

\subsection{The Open-Set Problem of Density Maps}
\begin{figure}[!ht]
	\begin{center}
		%\fbox{\rule{0pt}{1.8in} \rule{0.9\linewidth}{0pt}}
		\includegraphics[width=0.60\linewidth]{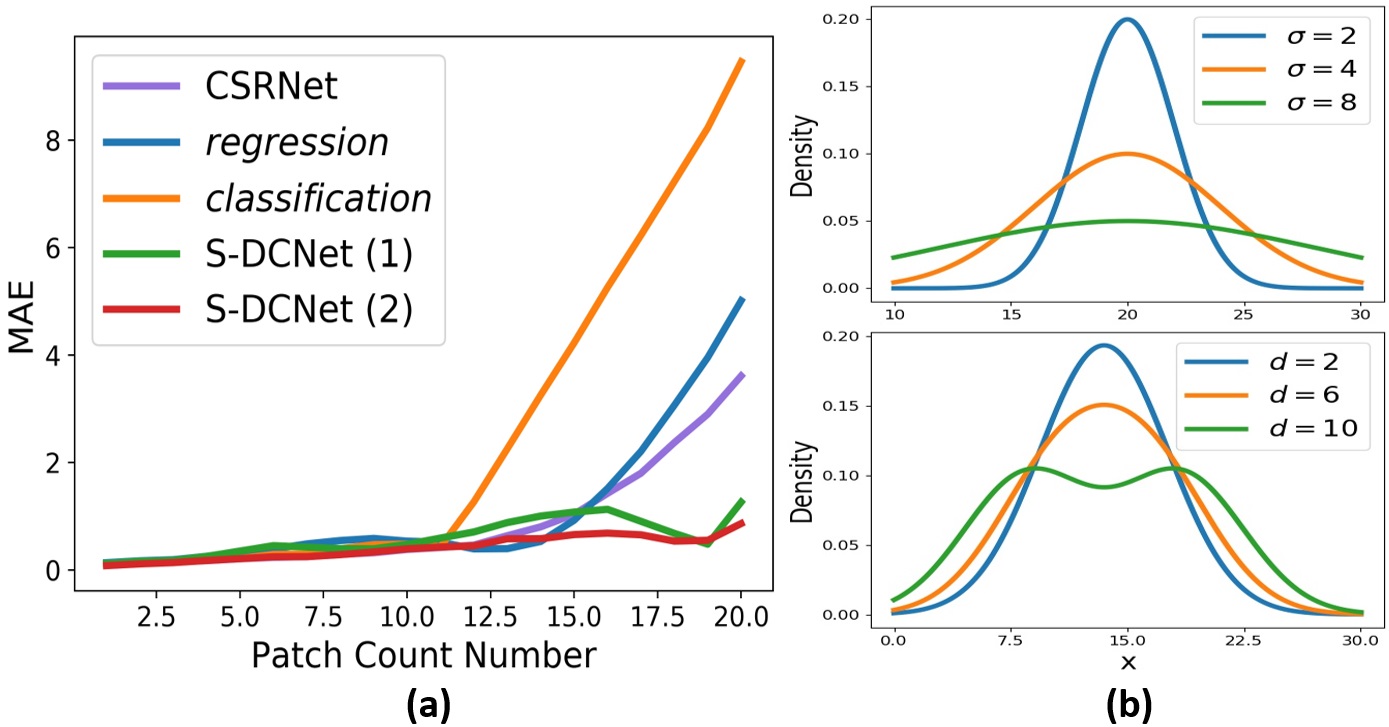}
	\end{center}
	\vspace{-10pt}
	\caption{(a) The toy-level experiment with an extra ``CSRNet'' baseline. (b) Density values along one axis with various kernels (top), and with two kernels with different relative distances.}
	\label{fig:toy_exp}
	\vspace{-15pt}
\end{figure}

Density maps are actually in the open set as well. As shown in Fig.~\ref{fig:toy_exp}(b) (top), for a single point, different kernel sizes lead to different density values. When multiple objects exist and are close, density patterns are even much diverse as in Fig.~\ref{fig:toy_exp}(b) (bottom). Since observed samples are limited, density maps are certainly in an open set.

We add another baseline of CSRNet~\cite{CSRNet_2018_CVPR} to the toy experiment in Fig.~\ref{fig:toy_exp}(a). CSRNet also performs worse than S-DCNet in the open set ($>10$), which implies the open-set problem also exists in density map based methods.

Furthermore, density map cannot be used in S-DCNet, because it is not spatially divisible. This is determined by its physical definition. However, local counts can. Thus we adopt local counts in S-DCNet rather than density maps.

\subsection{Relation to Other Methods}
\paragraph{IG-CNN~\cite{Divide_grow_2018_CVPR}}
IG-CNN drew inspirations from ensemble learning and trained a series of networks to tackle different scenes. While our S-DCNet focuses on inducing and utilizing physical laws, such as the “open set” problem in counting and the spatial divisibility of local counts. We propose to transform the open-set counting into a closed-set problem via spatial divide-and-conquer.  

\paragraph{Attention Mechanisms}
Despite it is possible to provide explicit supervision to $W_i$, we find that S-DCNet already can produce reasonably good divisions with the implicit supervision provided by $L_R^2$. This has another benefit, the network can learn when to divide not just in counts larger than $C_{max}$. The visualizations of $W_i$s in Fig.~\ref{fig:w_visual} further justify our point. %The direct supervision is a good suggestion, we will further investigate this in the extension of current work.
To highlight the difference against attention, we remove the division decider and generate a three-channel output conditioned on $F_2$, then process it with softmax to obtain $W_0^{att},W_1^{att},W_2^{att}$. The final count is merged as $W_0^{att}*upsample(C_0)+W_1^{att}*upsample(C_1)+W_2^{att}*C_2$. In SHTech PartA, it has $64.1$ $MAE$ and $109.9$ $MSE$ (worse than S-DCNet). As per the visualization of $W_i^{att}$ in Fig.~\ref{fig:w_visual}, we find the attention only focuses on the highest resolution and no effect of division is observed. In addition, S-DCNet executes fusion progressively, while attention fuses the prediction in a single step.

\begin{figure}[!ht]
	\begin{center}
		%\fbox{\rule{0pt}{1.8in} \rule{0.9\linewidth}{0pt}}
		\includegraphics[width=0.65\linewidth]{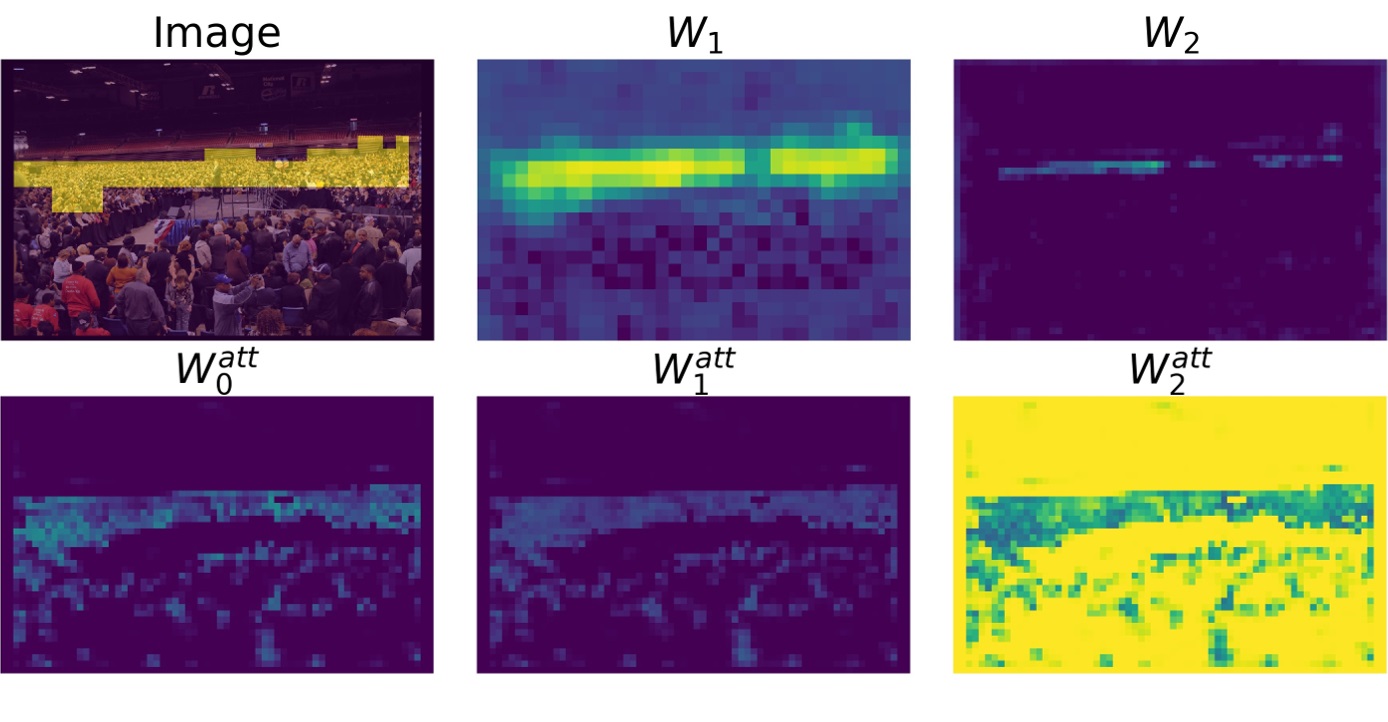}
	\end{center}
	\vspace{-15pt}
	\caption{Visualization of $W_i$ for S-DCNet (top) and the attention baseline (bottom). The lighter the image is, the greater the values are. In the input image, count values greater than $C_{max}$ are indicated by yellow regions.}
	\label{fig:w_visual}
	\vspace{-15pt}
\end{figure}

\subsection{Further Discussions on S-DCNet}
\paragraph{The necessity to distinguish counting task into open set and closed set scenarios}
One may raise the concern like: the relevance of distinguishing counting to an open set and closed set is unnecessary if each data point (head) is treated separately and the network learns to count each data point. If the network can count each head well, counting should already be addressed by detection networks. However, detection performs poorly when objects seriously overlap. This is why the notion of density map is introduced in~\cite{vlaz2010denlearn}, and density-based networks beat detection networks in counting. 
It is thus not suitable to treat each point separately, and distinguishing counting to an open set and closed set makes sense.
\paragraph{Generating ground-truth local counts}
Generating local counts directly from point annotations does not take partial objects cropped in patches into account. Density maps naturally tackle this situation. Thus we generate ground-truth counts of local patches by integrating over the density maps. This strategy is only utilized during training, while the point annotations are still used to calculate errors during validation.
\paragraph{If one position in $W_1$ is $0$, which means the initial prediction should not be replaced. Is it possible that the same position in $W_2$ is $1$?}
In theory, it is possible, because each division decision is independent. However, in practice, we do not observe such a behaviour of $W$ (Fig.~\ref{fig:w_visual}). Even this situation appears, we do not think it will be a problem. $W_2$ gives the second chance for division if the division decider makes a wrong decision in $W_1$.
\paragraph{Why $C_2$ is performing much worse than $C_1$ and $C_0$ in S-DCNet?}
$C_0$, $C_1$ and $C_2$ are trained jointly in S-DCNet and greatly influenced by the loss of $L_R^2$. As shown in Fig.~\ref{fig:w_visual}, $W_2$ focus on local patches with high density, which means $L_R^2$ will push $C_2$ to predict well on these patches and ignore others. High density patches, however, only occupy a small fraction. $C_2$ thus tends to predict worse than $C_0$ and $C_1$. This may also explain why three-stage/four-stage S-DCNet performs worse than two-stage S-DCNet.

%\newpage
\subsection{Qualitative Results of S-DCNet}
\begin{figure*}[th]
	\begin{center}
		\includegraphics[width=0.8\linewidth]{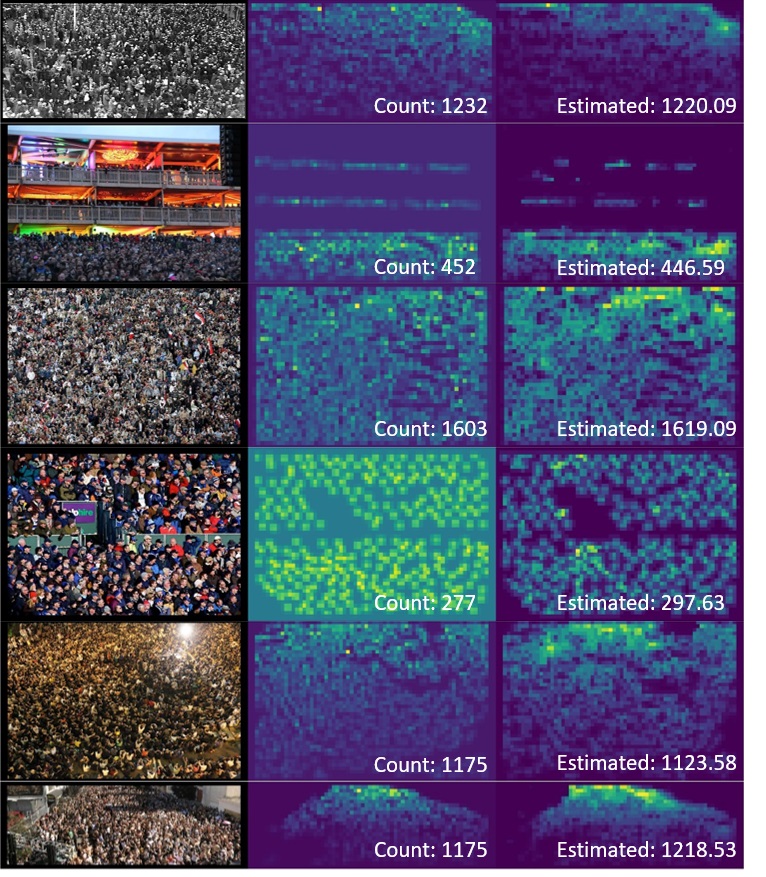}
	\end{center}
	%\vspace{-10pt}
	\caption{Some samples generated by S-DCNet from the test set of ShanghaiTech Part\_A dataset. The left column shows the original images, while the middle and right columns display the ground truth and predicted count maps respectively.}
	\label{fig:SHA_res}
	\vspace{-10pt}
\end{figure*}

\begin{figure*}[th]
	\begin{center}
		\includegraphics[width=0.8\linewidth]{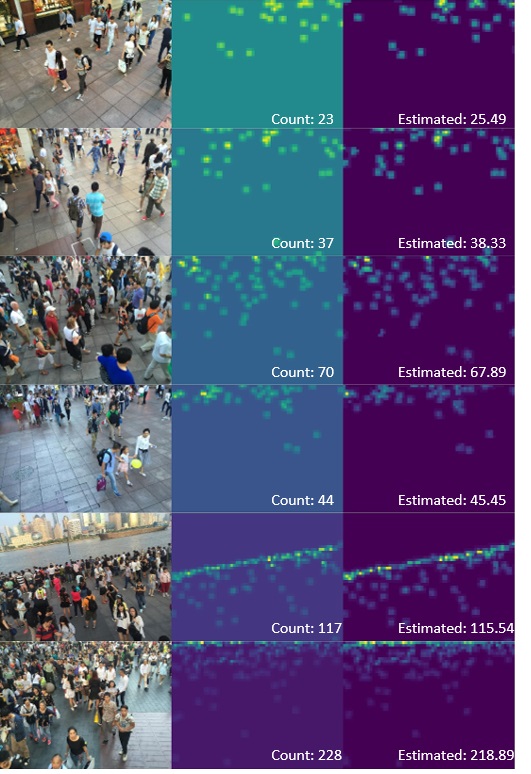}
	\end{center}
	%\vspace{-10pt}
	\caption{Some samples generated by S-DCNet from the test set of ShanghaiTech Part\_B dataset. The left column shows the original images, while the middle and right columns display the ground truth and predicted count maps respectively.}
	\label{fig:SHB_res}
	\vspace{-10pt}
\end{figure*}

\begin{figure*}[th]
	\begin{center}
		\includegraphics[width=0.8\linewidth]{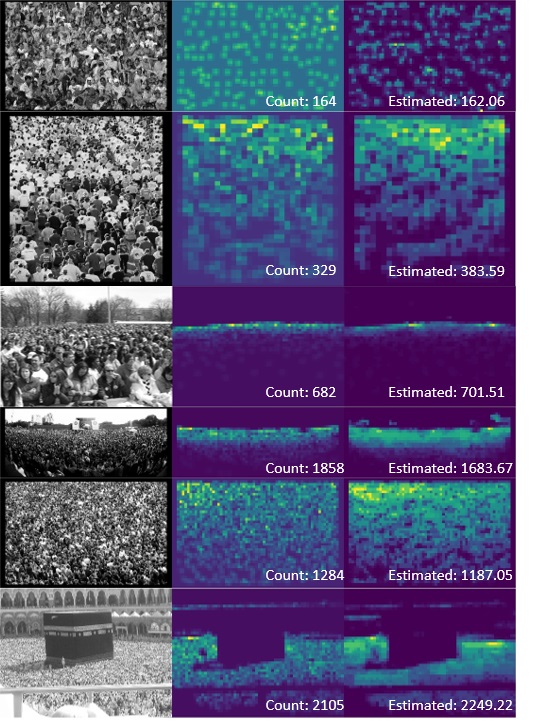}
	\end{center}
	%\vspace{-10pt}
	\caption{Some samples generated by S-DCNet from the test set of UCF\_CC\_50 dataset. The left column shows the original images, while the middle and right columns display the ground truth and predicted count maps respectively.}
	\label{fig:UCFCC50_res}
	\vspace{-10pt}
\end{figure*}

\begin{figure*}[th]
	\begin{center}
		\includegraphics[width=0.8\linewidth]{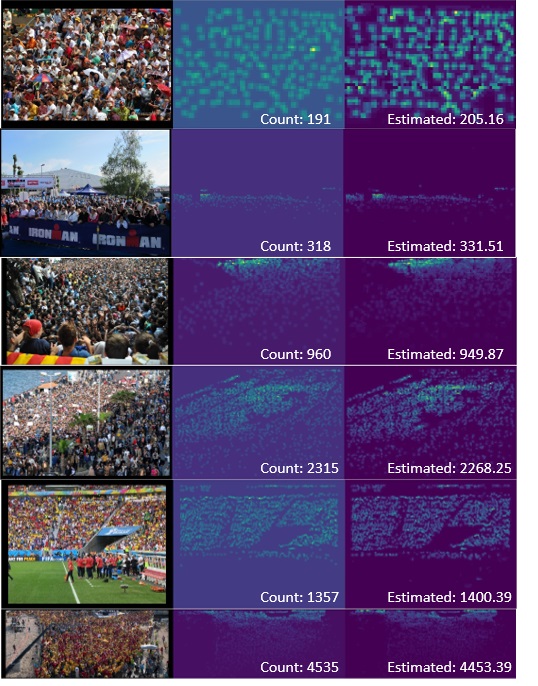}
	\end{center}
	%\vspace{-10pt}
	\caption{Some samples generated by S-DCNet from the test set of UCF-QNRF dataset. The left column shows the original images, while the middle and right columns display the ground truth and predicted count maps respectively.}
	\label{fig:UCQF_res}
	\vspace{-10pt}
\end{figure*}

\begin{figure*}[th]
	\begin{center}
		\includegraphics[width=0.8\linewidth]{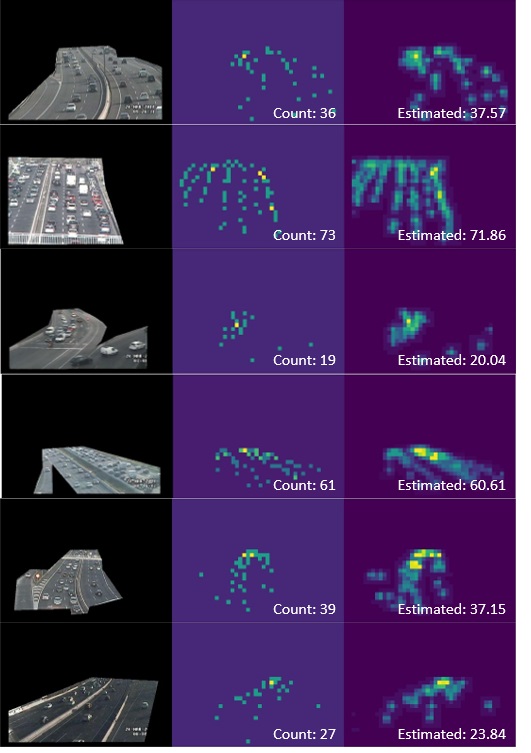}
	\end{center}
	%\vspace{-10pt}
	\caption{Some samples generated by S-DCNet from the test set of TRANCOS dataset. The left column shows the original images, while the middle and right columns display the ground truth and predicted count maps respectively.}
	\label{fig:TRANCOS_res}
	\vspace{-10pt}
\end{figure*}

\begin{figure*}[th]
	\begin{center}
		\includegraphics[width=0.8\linewidth]{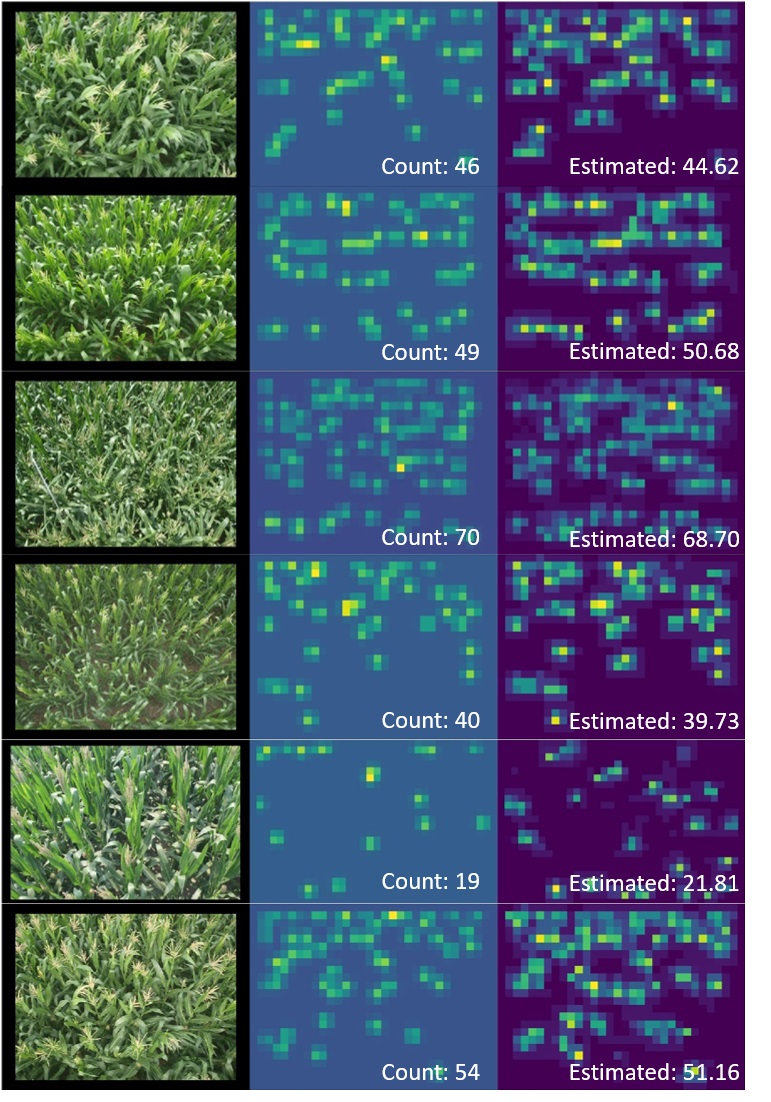}
	\end{center}
	%\vspace{-10pt}
	\caption{Some samples generated by S-DCNet from the test set of MTC dataset. The left column shows the original images, while the middle and right columns display the ground truth and predicted count maps respectively.}
	\label{fig:MTC_res}
	\vspace{-10pt}
\end{figure*}

We present some qualitative results of two-stage S-DCNet on five benchmarks (ShanghaiTech, UCF\_CC\_50, UCF-QNRF, TRANCOS and MTC) in Fig.~\ref{fig:SHA_res} to~\ref{fig:MTC_res}. S-DCNet predicts the local count map conditioned on the input image, where each element denotes a count value of the corresponding $16\times16$ local area. Meanwhile, since the output stride of S-DCNet is $64$, we pad the original image with zeros to ensure that the length and width are multiples of $64$.

% SHA

\end{document}